\documentclass{article}
\usepackage[preprint, nonatbib]{nips_2018}
\usepackage{float}
\usepackage[square,numbers,sort&compress]{natbib}
\usepackage{multirow}
\usepackage{algorithm}
\usepackage{algpseudocode}
\usepackage[utf8]{inputenc}
\usepackage[T1]{fontenc}
\usepackage{hyperref}
\usepackage{url}
\usepackage{booktabs}
\usepackage{amsfonts}
\usepackage{nicefrac}
\usepackage{microtype}
\usepackage{graphicx}
\usepackage{tabularx}
\usepackage{amsmath}
\usepackage{array}
\usepackage{graphicx}
\usepackage{indentfirst} 
\usepackage{titlesec}
\usepackage{caption}

\newcolumntype{Y}{>{\raggedright\arraybackslash}X}

\newcolumntype{L}{>{\raggedright\arraybackslash}X}

\DeclareUnicodeCharacter{2460}{\textcircled{\scriptsize{1}}}
\DeclareUnicodeCharacter{2461}{\textcircled{\scriptsize{2}}}
\DeclareUnicodeCharacter{2462}{\textcircled{\scriptsize{3}}}
\DeclareUnicodeCharacter{2463}{\textcircled{\scriptsize{4}}}
\DeclareUnicodeCharacter{2464}{\textcircled{\scriptsize{5}}}
\DeclareUnicodeCharacter{2465}{\textcircled{\scriptsize{6}}}
\DeclareUnicodeCharacter{2466}{\textcircled{\scriptsize{7}}}
\DeclareUnicodeCharacter{2467}{\textcircled{\scriptsize{8}}}

\title{Generating Comprehensive Lithium Battery Charging Data with Generative AI}

\author{\authors}

\author{%
  \parbox{\linewidth}{\centering%
  Lidang Jiang\textsuperscript{a}, Changyan Hu\textsuperscript{a}, Sibei Ji\textsuperscript{a}, Hang Zhao\textsuperscript{a}, Junxiong Chen\textsuperscript{b}*, Ge He\textsuperscript{a}* \\[1ex]
  \textsuperscript{a}School of Chemical Engineering, Sichuan University, Chengdu 610065 Sichuan, PR China \\[1ex]
  \textsuperscript{b}School of Automation Engineering, University of Electronic Science and Technology of China, Chengdu, 610023, China \\[1ex]
  \texttt{hege@scu.edu.cn, cjxzlm@my.swjtu.edu.cn}
  }
}

\titlespacing*{\section}
  {0pt} 
  {2\parskip} 
  {2\parskip} 

\titlespacing*{\subsection}
  {0pt} 
  {2\parskip} 
  {2\parskip} 

\titlespacing*{\subsubsection}
  {0pt} 
  {2\parskip} 
  {2\parskip} 

\begin{document}

\maketitle
\begin{abstract}
In optimizing performance and extending the lifespan of lithium batteries, accurate state prediction is pivotal. Traditional regression and classification methods have achieved some success in battery state prediction. However, the efficacy of these data-driven approaches heavily relies on the availability and quality of public datasets. Additionally, generating electrochemical data predominantly through battery experiments is a lengthy and costly process, making it challenging to acquire high-quality electrochemical data. This difficulty, coupled with data incompleteness, significantly impacts prediction accuracy. Addressing these challenges, this study introduces the End of Life (EOL) and Equivalent Cycle Life (ECL) as conditions for generative AI models. By integrating an embedding layer into the CVAE model, we developed the Refined Conditional Variational Autoencoder (RCVAE). Through preprocessing data into a quasi-video format, our study achieves an integrated synthesis of electrochemical data, including voltage, current, temperature, and charging capacity, which is then processed by the RCVAE model. Coupled with customized training and inference algorithms, this model can generate specific electrochemical data for EOL and ECL under supervised conditions. This method provides users with a comprehensive electrochemical dataset, pioneering a new research domain for the artificial synthesis of lithium battery data. Furthermore, based on the detailed synthetic data, various battery state indicators can be calculated, offering new perspectives and possibilities for lithium battery performance prediction.
\end{abstract}

\setlength{\parindent}{2em}

\section{ Introduction}

Lithium-ion batteries (LIBs) have emerged as a pivotal component in energy conversion strategies due to their low cost, high energy density, and longevity, poised to replace traditional fossil-fueled automotive engines. This plays a critical role in global efforts to reduce greenhouse gas emissions and combat climate change\cite{hofmann2016assessment}. As an environmentally friendly and efficient rechargeable energy option, lithium-ion batteries have found extensive applications in the realm of intelligent manufacturing, including computational engineering, logistics, and aerospace, among others. Their low pollution footprint, low self-discharge rates, broad operating temperature range, high energy and power density, along with long-term durability\cite{jiang2018low,opitz2017can,xiong2017double}, have positioned them as a preferred energy solution. However, despite their vast potential across various application domains, the high replacement costs\cite{kacetl2022design}, challenges in accurately assessing battery states\cite{zhang2022identification}, and safety concerns\cite{huang2016experimental,wang2012thermal,wang2021critical} continue to be focal points of widespread user attention. Fortunately, Recurrent Neural Networks (RNN)\cite{liu2010adaptive}, along with their derivatives\cite{bian2020stacked,ma2021state,shu2020stage,tian2021state}, and Convolutional Neural Networks (CNN) \cite{tian2021deep}, have been proven effective in precisely predicting battery states.

However, the efficacy of these data-driven methods largely hinges on the availability and quality of public datasets. With the growing interest in data-driven techniques and the pursuit of a deeper understanding of battery complex interactions, various datasets featuring different battery chemistries, quantities of tested batteries, and testing conditions have been developed. These datasets can generally be categorized into four types: cycle ageing data\cite{he2011prognostics,preger2020degradation,raj2020investigation,williard2013comparative}, drive cycle data\cite{kollmeyer2018panasonic}, calendar ageing data\cite{calce2018batteryData}, and specific-purpose datasets such as simulated satellite operation profile battery data\cite{cameron2015battery,kulkarni2008small}. Among these, cycle ageing datasets are the most common type in current practice, aiming to experimentally study the impact of various factors (e.g., charge current, discharge current, temperature, depth of discharge DOD) on the battery's capacity retention ability during cycling. These datasets typically include measurements of current, voltage, and temperature during the cycles, along with the capacity and internal resistance or impedance measurements for each cycle. The dataset released by NASA\cite{saha2007battery}, which covers information on 34 lithium-ion 18650 batteries with a nominal capacity of 2 Ah, marked the advent of the first publicly available battery dataset, significantly impacting the battery research field. The batteries underwent cycling tests at different environmental temperatures (4°C, 24°C, 43°C) using a standard constant current-constant voltage (CC-CV) charging protocol, and various discharge regimes were implemented. Another vital resource is the collaboration between the Toyota Research Institute (TRI), MIT, and Stanford University, which launched two rich and user-friendly high-throughput cycling test datasets involving 357 (124+233) commercial Lithium Iron Phosphate (LFP)/Graphite batteries (APR18650M1A) produced by A123 Systems, with a rated capacity of 1.1 Ah. The dataset of 124 batteries\cite{severson2019data} aims to study the effects of fast charging protocols on battery ageing, while the dataset of 233 batteries\cite{attia2020closed} is similar to the former and is not repeated here. Particularly, the dataset of 124 batteries, due to its high-quality data, well-organized format, and accessibility, made a significant impact on the entire field upon its release in 2019.

Creating these battery datasets requires significant time and financial resources, yet the challenge of acquiring high-quality electrochemical data remains unresolved. Introducing synthetic data can enhance existing datasets, thereby improving model training performance. A popular method currently involves interpolating experimental data\cite{pyne2019generation}. Although this strategy is straightforward, the generated data are limited in quality and diversity. On the other hand, artificial intelligence generative models offer new possibilities for data generation. Research on generative AI has mainly focused on natural language processing (NLP) and computer vision (CV) fields. In NLP, the introduction of the Transformer model\cite{vaswani2017attention} in 2017 further propelled advancements in this domain. Transformers process input sequences through an encoder-decoder architecture, generating hidden representations and output sequences. Its key innovation—multi-head attention mechanism—allows the model to allocate different attention weights based on the relevance between words, significantly enhancing the model's ability to handle long-term dependencies. Additionally, the architectural characteristics of Transformers enable high parallel processing capabilities, achieving exceptional performance improvements across various NLP tasks. In the CV domain, the introduction of Generative Adversarial Networks (GANs)\cite{goodfellow2014generative} in 2014 marked a significant turning point. However, the adversarial training mechanism between the generator and discriminator makes finding the optimal balance highly challenging. Subsequent research efforts have focused on improving model generalization capabilities and the quality of generated data through structural innovations (e.g.,  StyleGAN\cite{karras2019style}) and loss functions (e.g., WGAN\cite{gulrajani2017improved}). Inspired by the Transformer model, researchers developed the Vision Transformer (ViT)\cite{dosovitskiy2020image}, enabling efficient processing of image-related downstream tasks. More recently, diffusion generative models\cite{song2019generative} have emerged as a cutting-edge technology for generating high-quality images. They generate data by gradually introducing and reversing multilevel noise perturbations, a process of progressive degradation and repair that endows diffusion models with powerful image generation capabilities.

With the rapid advancement of computational hardware such as GPUs, distributed training, and cloud computing technologies, large-scale models based on fundamental units like Transformer, ViT, and diffusion models, including BERT\cite{devlin2018bert}, GPT\cite{radford2018improving}, and stable diffusion\cite{rombach2022high}, have emerged like mushrooms after the rain. However, due to the Transformer model's demand for GPU memory being directly proportional to the square of the sequence length, generative models based on Transformer require relatively high hardware specifications. Meanwhile, although diffusion models adopt a progressive strategy during generation, their efficiency is relatively low. Variational Autoencoders (VAE)\cite{kingma2013auto} attempt to generate data close to the original input by mapping data to a probability distribution and learning its reconstruction. On this basis, the introduction of Conditional Variational Autoencoders (CVAE)\cite{sohn2015learning} has made it possible to generate targeted and supervised images or sequence samples. Compared to the training challenges of GANs and the high hardware requirements of Transformer-based models, CVAEs, due to their effectiveness in precisely modeling data distributions and ease of training, have been widely used across various application scenarios. Additionally, embedding layers\cite{mikolov2013efficient} play a central role in handling categorical data within neural networks, especially indispensable in the process of encoding vocabulary in NLP tasks. Embedding layers can map various types of categorical data (such as words, characters, or different types of labels) to dense vectors (i.e., embedding vectors) in high-dimensional space, effectively revealing complex relationships between categories, such as semantic similarity.

In this study, we utilized End of Life (EOL) and Equivalent Cycle Life (ECL) as conditions for a generative AI model, incorporating an embedding layer into the CVAE framework to develop the Refined Conditional Variational Autoencoder (RCVAE). Leveraging the authoritative MIT dataset, we preprocessed data into a quasi-video format to achieve a comprehensive integration of electrochemical data such as voltage, current, temperature, and charging capacity, which was then processed through the RCVAE model. Supported by customized training and inference algorithms, our model is capable of generating detailed and high-quality electrochemical charging data for EOL and ECL under supervised conditions. This approach provides a complete electrochemical dataset, paving a new pathway for the artificial synthesis of lithium battery data. Moreover, the detailed synthetic data enables the calculation of various battery state indicators, offering fresh perspectives and possibilities for lithium battery performance prediction.

\section{Methods}

\begin{figure}[H]
\centering
\includegraphics[width=\textwidth]{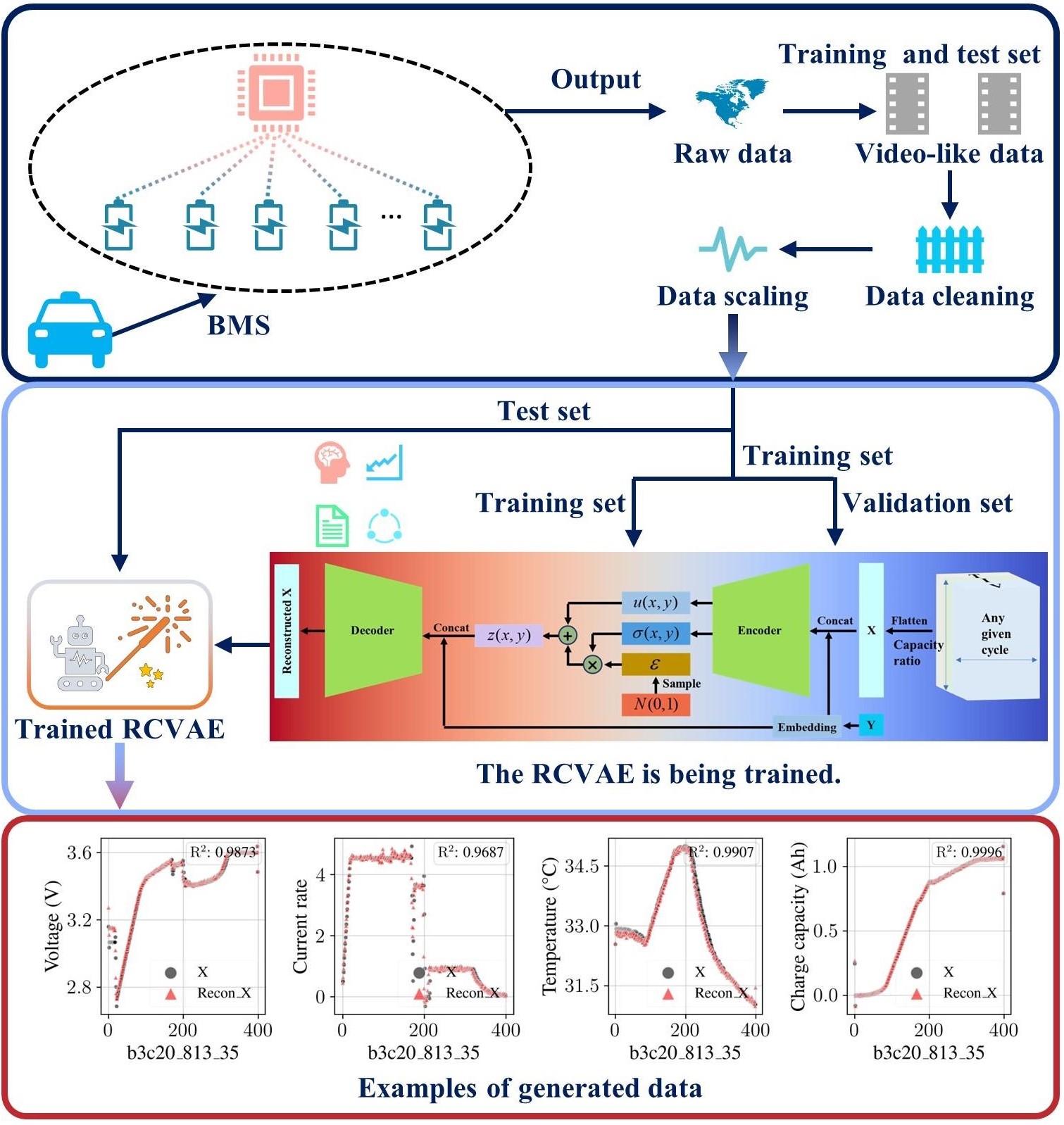}
\caption{ illustrates the technical roadmap for generating high-quality electrochemical data with RCVAE. }
\label{fig5}
\end{figure}

After a detailed introduction of the dataset, this section will focus on elucidating the complete workflow of RCVAE in generating electrochemical data. As illustrated in Figure~\ref{fig5}, the workflow begins with the Battery Management System (BMS), which is responsible for collecting real-time operational data from the batteries. These raw data are then transformed into a quasi-video format through a series of preprocessing steps to enhance data processability. The preprocessed data are subsequently fed into the RCVAE model for training and to perform prediction tasks.

\subsection{Data generation}

\begin{figure}[H]
\centering
\includegraphics[width=\textwidth]{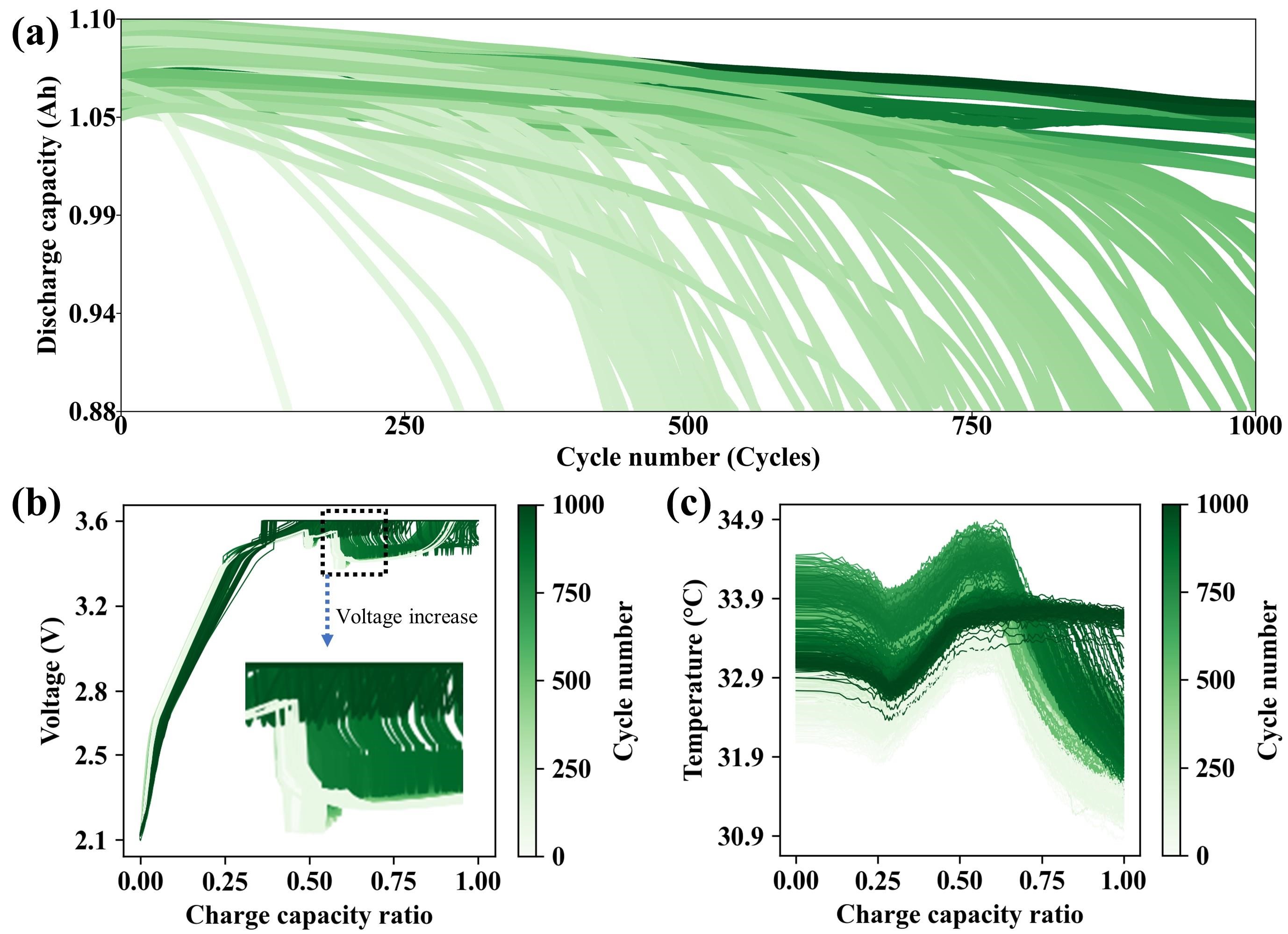}
\caption{conducts a comprehensive analysis of lithium-ion battery performance: (a) based on the MIT dataset, showing the trend of lithium-ion battery discharge capacity decay over cycles; (b) displaying the variation in voltage of the "b3c0" battery across different charging cycles, with the voltage decline areas highlighted by black square markers, emphasizing the voltage decay characteristics during charging; (c) describing the temperature change trend of the "b3c0" battery during charging, reflecting the thermal management status at different charging stages. }
\label{fig6}
\end{figure}

The dataset used in this study was created by Severson et al. in 2019\cite{severson2019data}, known as the MIT dataset, which has become one of the most authoritative data sources in this field. This battery dataset includes 124 batteries, achieving a diversity of battery lifespans by controlling different charging and discharging current sizes. As shown in Figure~\ref{fig6}a, the selected batteries are lithium iron phosphate/graphite batteries produced by A123 Systems, with a nominal capacity of 1.1Ah. The cycle number at which the battery's discharge capacity falls to 80\% of its nominal capacity is defined as the EOL for the battery. This design effectively reflects the variability in battery lifespan observed in the real world.

The dataset comprises three batches of batteries. Figure~\ref{fig6}b displays the variation in charging voltage for the first battery (identified as ``b3c0'') in the third batch. Battery ``b3c0'' is chosen as an example because it exhibits fewer outliers than others, facilitating graphical presentation. As depicted in Figure~\ref{fig6}b, the battery voltage changes with increasing cycle numbers, reflecting the degradation of battery performance. Even for the same battery, the voltage data across different cycles vary with the number of cycles, where the cycle count to some extent indicates the degree of battery degradation. The variance in voltage data across different cycles suggests that voltage data from any cycle can serve as unique features mapping to EOL. Similarly, in Figure~\ref{fig6}c, a comparable conclusion can be drawn: temperature data across different cycles for the same battery show variability, indicating that temperature data from any cycle can also serve as distinct features mapping to EOL.

\subsection{Data Preprocessing}

\begin{figure}[H]
\centering
\includegraphics[width=\textwidth]{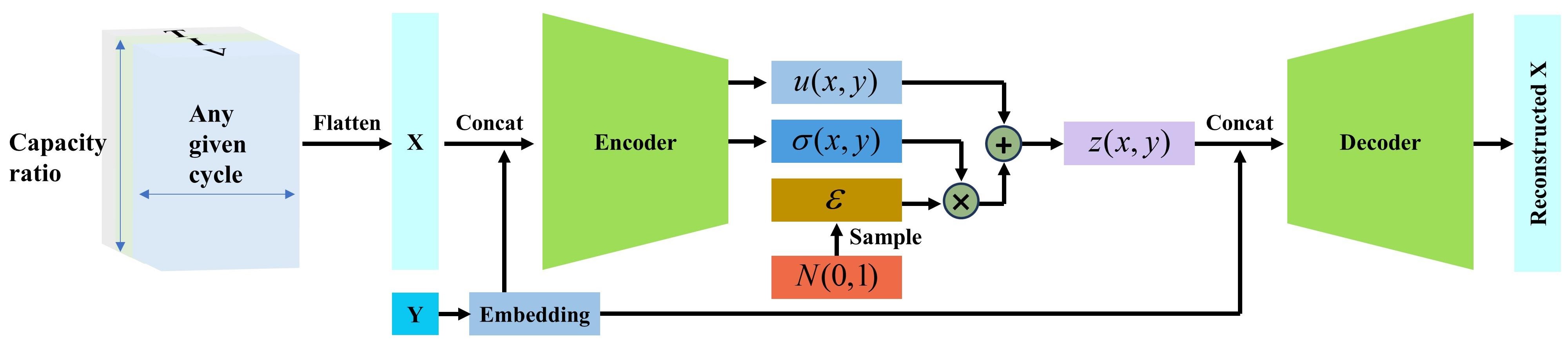}
\caption{describes the quasi-video data after preprocessing and the overall architecture of RCVAE. }
\label{fig7}
\end{figure}

For data preprocessing, this study draws on the methodology described in the literature\cite{yang2021machine}. After organizing the samples in a quasi-video format, each of the 124 batteries generated n samples, resulting in a total of 124n samples. To reduce the randomness in experimental outcomes, this study shuffles the data at the sample level rather than at the battery level, meaning samples from the same battery will not appear consecutively in the dataset. Following the proportion used in other research\cite{yang2021machine,zhang2022deep}, the first 94n samples were divided into a training set, with the remaining samples allocated to a test set. Subsequently, data cleaning and scaling were performed. The training set was then further divided into a new training set and a validation set at a ratio of 4:1. Unlike previous studies that utilized data from the first 10 cycles, this study used data from any cycle as a sample, as depicted in Figure~\ref{fig7}. After preprocessing, the data was transformed into a five-dimensional array processable by 3D CNNs. The first dimension represents the number of samples, the second dimension represents the number of channels, indicating three types of data (voltage, current, temperature), and the third dimension represents depth, a dimension added due to the inclusion of charging capacity data. The number of data points for a single type of data forms another dimension, which is reshaped into two dimensions and placed in the height and width directions, forming a five-dimensional electrochemical feature array, as shown on the left side of Figure~\ref{fig7}.

\subsection{RCVAE (Refined Conditional Variational AutoEncoders ) }
\subsubsection{Generating Embedding Vectors Through Labels}

Before the data is input into RCVAE, all dimensions other than the number of samples are flattened into sequences. Initially, the data's labels are processed through an embedding layer. As demonstrated in Figure~\ref{fig6}, data with different EOLs exhibit distinct characteristics, and similarly, data with different Remaining Useful Lives (RUL) also show their unique differences. Furthermore, if two batteries have similar lifespans, their RUL and electrochemical data will also display similarities. The calculation of RUL follows the formula(\ref{eq1}), where EOL represents the expected lifespan of the battery, and ECL refers to the number of cycles the battery has already completed. RUL is the difference between EOL and ECL:
\begin{equation} \label{eq1} 
\text{RUL} = \text{EOL} - \text{ECL}
\end{equation} 
Thus, through EOL and RUL, we can delineate the characteristics of electrochemical data; in this study, "EOL\_ECL" is used as a condition in RCVAE, serving as the label for this research.

Different "EOL\_ECL" combinations form a condition set $C=\left[c_1, c_2, \ldots, c_N\right]$, where each $c_n$ represents the $n$th unique category label $y$ in the sequence. Before $C$ enters the embedding layer, since "EOL\_ECL" exists in string format, it needs to be converted into integer indices for numerical computation. The mapping from $C$ to integer indices is achieved through a lookup table, which can be considered a function mapping category labels to unique integer indices. This mapping function can be represented as $f: C \rightarrow\{0,1,2, \ldots, N-1\}$, where $N$ is the total number of categories. For any category $c \in C$, its integer index can be represented as:

\begin{equation} \label{eq2} 
i=f(c)
\end{equation} 
Here, $c$ is the specific category label. $f$ is the mapping function from category labels to integer indices. $i$ is the integer index corresponding to $c$. Therefore, the integer index sequence converted through the mapping function $f$ is represented as:

\begin{equation} \label{eq3} 
I=\left[f\left(c_1\right), f\left(c_2\right), \ldots, f\left(c_N\right)\right]
\end{equation} 
Then, the integer indices $I=\left[i_1, i_2, \ldots, i_N\right]$ are passed into the embedding layer, which is also a lookup operation, mapping discrete category labels into a continuous vector space. Suppose $E$ is the embedding matrix, $i$ is the integer index of the category label, $v$ is the vector mapped through the embedding layer, then the embedding operation can be represented as:
\begin{equation} \label{eq4} 
v=E[i]
\end{equation} 
Here, the size of $E$ is $N \times D: N$ is the total number of categories, i.e., the number of elements in the $C$ set. $D$ is the dimension of the embedding vector (embedding\_dim). $E[i]$ represents the row in the embedding matrix $E$ indexed by $i$, i.e., the embedding vector corresponding to index $i$. Thus:

\begin{equation} \label{eq5} 
V=E\left[i_1, i_2, \ldots, i_N\right]=\left[E\left[i_1\right], E\left[i_2\right], \ldots, E\left[i_N\right]\right]
\end{equation} 
\subsubsection{Forward Propagation of RCVAE}
The flow of preprocessed data within RCVAE is divided into several steps. Input: Initially, the embedding vectors $V$ (representing processed labels or other categorical information) and the feature data $X$ (the actual features of the input data) are concatenated to form a combined feature vector. This vector is then jointly input into the encoder. The dimensions of the embedding vector $V$ and the feature data $X$ are $d_V$ and $d_X$, respectively. The concatenation process can be represented as:
\begin{equation} \label{eq6} 
X_{\text {combined }}=\operatorname{concat}(V, X)
\end{equation} 
Here, concat $(\cdot$,$) denotes the concatenation operation, aligning V$ and $X$ in sequence to form a new combined feature vector with dimensions $d_V+d_X$. This combined feature vector $X_{\text {combined }}$ subsequently serves as the input for the encoder. The purpose of this step is to merge the label or category information related to the input data (represented by the embedding vector $V$ ) with the actual feature data $X$, to consider both aspects of information in the ensuing encoding process. Thus, integrating the embedding vector and feature data into a unified input lays the groundwork for subsequent encoder processing.

The encoder learns the distribution parameters (namely, mean $\mu$ and variance $\sigma^2$ ) of the latent representation of the input data $x$ and associated condition $y$. This can be expressed as the mean $\mu(x, y)$ and variance $\sigma^2(x, y)$ :
\begin{align}
\mu(x, y) & =f_\mu\left(X_{\text {combined }}\right) \label{eq7} \\
\log \left(\sigma^2(x, y)\right) & =f_\sigma\left(X_{\text {combined }}\right) \label{eq8}
\end{align}

Functions $f_\mu$ and $f_\sigma$ are implemented by multi-layer fully connected neural networks, designed to extract the mean and the logarithm of the variance from the input data. Here, the number of layers in the multi-layer fully connected neural network (MLP) is set as a hyperparameter and is not fixed. In the MLP, each layer applies a linear transformation and a nonlinear activation function to the output of the previous layer. Let $L$ be the total number of layers in the network; for each layer $l=1,2, \ldots, L$, linear transformations and nonlinear activations can be represented as follows:
\begin{align}
h_l & =W_l \cdot a_{l-1}+b_l \label{eq9} \\
a_l & =f\left(h_l\right) \label{eq10}
\end{align}

$W_l$ and $b_l$ are the weight and bias of the $l$ th layer, respectively, and $f$ is the nonlinear activation function. For the first layer, $a_0=X_{\text {combined }}$, meaning the combined feature vector serves as the input to the network. Ultimately, the mappings for the mean $\mu(x, y)$ and the logarithm of the variance $\log \left(\sigma^2(x, y)\right)$ can be represented by the output of the last layer. Assuming that the calculations for the mean and the logarithm of the variance are derived from the network's last two output layers respectively, then the calculation for the mean $\mu(x, y)$ can be represented as: $\mu(x, y)=W_\mu \cdot a_L+b_\mu$. The calculation for the logarithm of the variance $\log \left(\sigma^2(x, y)\right)$ can be represented as: $\log \left(\sigma^2(x, y)\right)=W_\sigma \cdot a_L+b_\sigma$. Here $W_\mu$ and $b_\mu$, as well as $W_\sigma$ and $b_\sigma$, are the weights and biases of the last two output layers specifically used for calculating the mean and the logarithm of the variance, with $a_L$ being the activation output of the last hidden layer. By this means, setting an MLP as a network with any number of layers allows for flexible adaptation to different complexities in model design requirements.

Next is the sampling of the latent representation: After obtaining the mean $\mu(x, y)$ and variance $\sigma^2(x, y)$, the latent representation $z(x, y)$ can be sampled from the corresponding Gaussian distribution using the reparameterization trick:

\begin{equation} \label{eq11} 
z(x, y)=\mu(x, y)+\epsilon \cdot \sqrt{\exp \left(\log \left(\sigma^2(x, y)\right)\right)}=\mu(x, y)+\sigma(x, y) \cdot \epsilon
\end{equation} 
where $\epsilon$ is noise sampled from the standard normal distribution $\mathcal{N}(0, I)$. This method allows the encoder not only to output the mean and variance of the latent representation of the input data under specific conditions but also to generate diverse outputs by introducing random noise $\epsilon$, a key characteristic of variational autoencoder generative models. In summary, the encoder operation learns to map the combination of input features and condition information to the distribution parameters of the latent space, generating the latent representation $z(x, y)$ through these parameters plus random noise. This process provides the foundation for the subsequent decoder to reconstruct the input or generate new data.

Preparation of Decoder Input: Before the latent variable $z(x, y)$ is input into the decoder, it needs to be concatenated with the corresponding embedding vector $V(y)$ to form the decoder's input vector. This can be represented by the following equation:

\begin{equation} \label{eq12} 
X_{\text {decoder }}=\operatorname{concat}(z(x, y), V(y))
\end{equation} 

By concatenating the latent variable $z(x, y)$ and the embedding vector $V(y)$ in sequence, a new vector $X_{\text {decoder }}$ is formed, which is then used as the input to the decoder. In this way, the decoder not only receives the latent representation mapped from the input data $x$ and condition $y$ but also integrates additional information about the condition $y$, enabling the decoder to generate target samples matching the input data under supervision while considering the condition information $y$. This combination of latent representation and condition information is one of the key features distinguishing CVAE from standard VAE.

Decoder Operation: The concatenated vector $X_{\text {decoder }}$ (a fusion of latent representation and label information) is inputted, and the target samples matching the input data are generated. It can be represented as:

\begin{equation} \label{eq13} 
X_{\text {reconstructed }}=g\left(X_{\text {decoder }}\right)
\end{equation} 
Here: $g(\cdot)$ is the function of the decoder, implemented through an MLP, aiming to learn how to reconstruct data from the latent space. This function is similar to that of the encoder and will not be reiterated here. $X_{\text {reconstructed }}$ is the reconstructed data output by the decoder, intended to resemble the original input data $x$ as closely as possible, while considering the condition information $y$. The learning objective of the decoder is to maximize the data likelihood given the latent representation and condition information, i.e., to maximize $p(x \mid z(x, y), V(y))$. By optimizing this objective, the decoder can generate target samples that match the original input data and consider the condition information. Specifically for the application of generating electrochemical samples, this means the decoder has learned how to generate new electrochemical samples that meet these conditions based on given electrochemical conditions and corresponding feature data. The forward propagation algorithm of RCVAE is shown as Algorithm ~\ref{algo1}.

\begin{algorithm}
\caption{Forward Propagation Algorithm of RCVAE}
\label{algo1}
\begin{algorithmic}
\State \textbf{Input:} Feature data $x$ and label $y$.
\State \textbf{Output:} Generated electrochemical sample by RCVAE.
\State \textbf{Step 1:} Embed label $y$ into $V(y)$ by Eq.(\ref{eq5}).
\State \textbf{Step 2:} Concat $x$ and $V(y)$ into one vector by Eq.(\ref{eq6}).
\State \textbf{Step 3:} Encode to latent space, get mean and variance by Eqs(\ref{eq7})-(\ref{eq8}).
\State \textbf{Step 4:} Sample latent variable $z(x,y)$ using mean and variance by Eq.(\ref{eq11}).
\State \textbf{Step 5:} Concat $z(x,y)$ and $V(y)$ for decoder by Eq.(\ref{eq12}).
\State \textbf{Step 6:} Decoder generates sample by Eq.(\ref{eq13}).
\State \textbf{Return:} Generated electrochemical sample by RCVAE.
\end{algorithmic}
\end{algorithm}

\subsubsection{Training and Evaluation of algorithm}

It's important to note that the embedding space for labels is determined based on the labels in the training set. However, labels in the test set may not directly correspond to the embedding space of the training set. For such test samples, their labels undergo preliminary preprocessing through a similarity strategy. Specifically, the similarity of test set labels is calculated using the difference formula(\ref{eq14}), by iterating through each label in the training set, calculating the distance to the test label, and then selecting the label with the smallest distance as the new test label. Subsequent processing steps are similar to those described earlier and are not detailed here.
\vspace{50pt}
\begin{equation} \label{eq14} 
distance=weight \cdot\left|\mathrm{EOL}_1-\mathrm{EOL}_2\right|+(1 -weight) \cdot\left|\mathrm{ECL}_1-\mathrm{ECL}_2\right|
\end{equation} 
The process of training and inferring RCVAE, as shown in Algorithm ~\ref{algo2}, starts with the hyperparameter tuning stage, where the best hyperparameters are selected based on the performance on the validation set using the Bayesian optimization algorithm. During the model training phase, to deeply learn the sample data, the validation set is merged with the training set to form a new training set on which the model is trained. The total loss during the training process is obtained by combining the Mean Squared Error (MSE(\ref{eq15})) and the Kullback-Leibler Divergence (KLD(\ref{eq16})), as shown in equation (\ref{eq17}).
\begin{equation} \label{eq15} 
\operatorname{MSE}=\frac{1}{K} \sum_{i=1}^K\left(x_i-\hat{x}_i\right)^2
\end{equation} 
Here, $K$ is the total number of samples in the batch, $x_i$ is the $i$ th element of the original input data, and $\hat{x}_i$ is the $i$ th element of the reconstructed data.
\begin{equation} \label{eq16} 
\mathrm{KLD}=-\frac{1}{2} \sum_{j=1}^J\left(1+\log \left(\sigma_j^2\right)-\mu_j^2-\sigma_j^2\right)
\end{equation} 
Where $J$ is the dimension of the latent space, $\mu_j$ and $\sigma_j^2$ are the mean and variance of the $j$ th latent variable, respectively.
\begin{equation} \label{eq17} 
\text { Loss }=\mathrm{MSE}+\frac{\mathrm{KLD}}{K}
\end{equation} 
In this part, normalizing by K ensures that the loss does not change with the size of the sample set, making training results comparable across different sample sizes. To ensure the model is thoroughly trained while remaining efficient, 1000 training epochs are set, with an early stopping mechanism introduced, and the early stopping counter set to 100. This means that if the loss(\ref{eq18}) on the validation set does not decrease for 100 consecutive training epochs, training will be terminated. During the model performance testing phase, the model's predictive performance is assessed for the first time using test set data. To comprehensively evaluate the model's predictive capability, this study uses the MAE and RMSE as evaluation metrics, with their respective mathematical expressions defined as equations(\ref{eq18}) and (\ref{eq19}).

\begin{align}
\text{MAE} & =\frac{1}{K} \sum_{i=1}^K\left|x_i-\hat{x}_i\right| \label{eq18} \\
\text{RMSE} & =\sqrt{\frac{1}{K} \sum_{i=1}^K\left(x_i-\hat{x}_i\right)^2} \label{eq19}
\end{align}

\noindent Where $\mathrm{K}$ is the total number of samples. $x_i$ is the ith element of the original data. $\hat{x}_i$ is the ith element of the reconstructed data.
\begin{algorithm}[H]
\centering
\caption{Training and Evaluation of Algorithm}
\label{algo2}
\begin{algorithmic}
\State \textbf{Input:} The training set, validation set, and test set.
\State \textbf{Output:} Trained Model, Test Loss
\State \textbf{Step 1:} Optimize hyperparameters using Bayesian optimization.
\State \textbf{Step 2:} Concatenate the training and validation sets to form a new training set.
\State \textbf{Step 3:} Set the hyperparameters found to be optimal in the previous steps.
\State \textbf{Do}
\State \hspace{8mm} Counter = 0.
\State \hspace{8mm} \textbf{Repeat}
\State \hspace{16mm} Outer Loop for multiple epochs:
\State \hspace{16mm} \textbf{Do}
\State \hspace{24mm} \textbf{Forward Propagation:}
\State \hspace{24mm} \textbf{Step 4:} Compute training loss by Eq.(\ref{eq17}) and validation loss by Eq.(\ref{eq18}).
\State \hspace{24mm} \textbf{Backward Propagation:}
\State \hspace{32mm} \textbf{Step 5:} Calculate gradients based on the training loss.
\State \hspace{32mm} \textbf{Step 6:} Update network parameters.
\State \hspace{24mm} \textbf{Step 7:} If validation loss decreases, save the model and Counter = 0.
\State \hspace{44mm} Otherwise, Counter = Counter + 1.
\State \hspace{24mm} \textbf{Step 8:} If Counter exceeds a predefined threshold, trigger early stopping.
\State \hspace{16mm} \textbf{End}
\State \hspace{8mm} \textbf{Until} reaching a predefined number of epochs or early stopping is triggered.
\State \textbf{End}
\State \textbf{Step 9:} Save the model, Load the model.
\State \textbf{Step 10:} Forward Propagate on the test set to compute the test loss by Eq.(\ref{eq18}) - Eq.(\ref{eq19}).
\State \Return Trained Model, Test Loss
\end{algorithmic}
\end{algorithm}
\section{Results}
\subsection{Generative Results of RCVAE}

This study aims to explore the impact of dataset size on the RCVAE model's performance in generating electrochemical data. By analyzing the model's behavior with different magnitudes of early-cycle data, we observed some interesting trends. Due to the values of some data types approaching zero in certain cases, the Mean Absolute Percentage Error (MAPE) exhibited considerable instability, suggesting that MAPE might not be an appropriate metric for evaluation. As shown in Table~\ref{tab:tab1}, for voltage prediction, the Mean Absolute Error (MAE) fluctuated within the range of 0.022 to 0.023V, and the Root Mean Square Error (RMSE) ranged between 0.418 and 0.447V; in predicting the current rate, the MAE values were between 0.272 and 0.286, and the RMSE values varied from 0.686 to 0.755, demonstrating a consistently high accuracy; regarding temperature predictions, MAE values were between 0.404 and 0.417℃, with RMSE values ranging from 0.753 to 0.789℃, both indicating the model's high precision; the predictions for charging capacity were also encouraging, with MAE values between 0.019 and 0.021 Ah, and RMSE values from 0.039 to 0.044 Ah, showcasing the model's precise capture of battery charging capacity changes. To comprehensively assess the model's performance in generating voltage, current rate, temperature, and charging capacity data, the error metrics for these four types of data were considered together, resulting in a total MAE value between 0.126 and 0.130, and RMSE values maintained between 0.418 and 0.447.

\begin{table}[H]
\caption{Results of Electrochemical Data Generated by RCVAE}
\smallskip
\label{tab:tab1}
\begin{tabularx}{\textwidth}{YYYYYYY}
\toprule
\textbf{Data type} & \textbf{Error type} & \textbf{20 (Cycles)} & \textbf{40 (Cycles)} & \textbf{60 (Cycles)} & \textbf{80 (Cycles)} & \textbf{100 (Cycles)} \\
\midrule
\multirow{2}{*}{V (V)} & MAE & 0.023 & 0.023 & 0.022 & 0.022 & 0.022 \\
 & RMSE & 0.044 & 0.046 & 0.046 & 0.047 & 0.046 \\
\multirow{2}{*}{I} & MAE & 0.286 & 0.283 & 0.282 & 0.273 & 0.272 \\
 & RMSE & 0.686 & 0.731 & 0.748 & 0.755 & 0.737 \\
\multirow{2}{*}{T($^\circ$C)} & MAE & 0.404 & 0.41 & 0.417 & 0.405 & 0.404 \\
 & RMSE & 0.753 & 0.769 & 0.789 & 0.787 & 0.773 \\
\multirow{2}{*}{Qc (Ah)} & MAE & 0.021 & 0.021 & 0.02 & 0.019 & 0.02 \\
 & RMSE & 0.039 & 0.042 & 0.044 & 0.043 & 0.044 \\
\multirow{2}{*}{Total} & MAE & 0.129 & 0.13 & 0.13 & 0.126 & 0.126 \\
 & RMSE & 0.418 & 0.436 & 0.446 & 0.447 & 0.439 \\
\bottomrule
\end{tabularx}
\end{table}
\unskip

\begin{figure}[H]
\centering
\includegraphics[width=\textwidth]{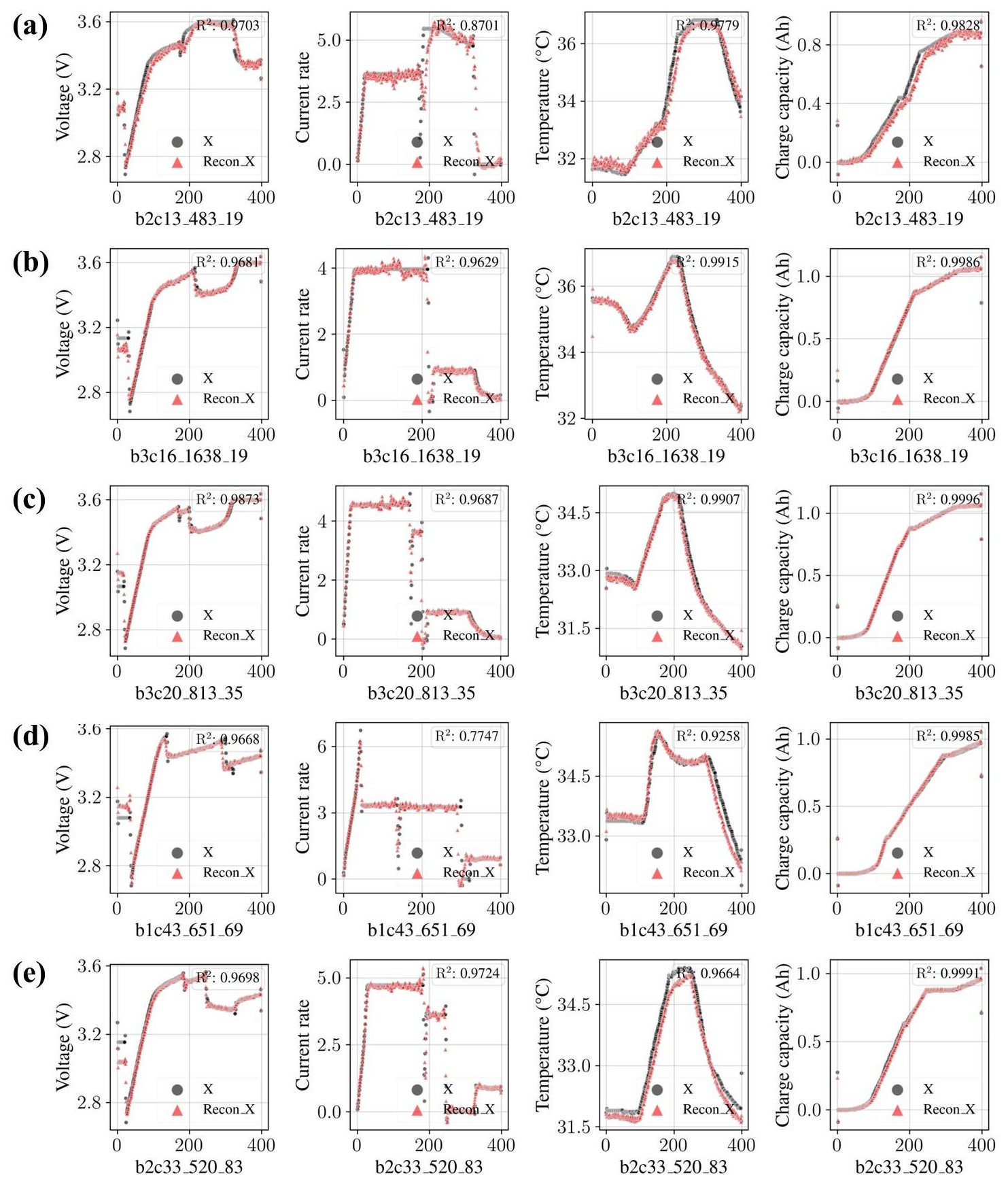} 
\caption{displays the voltage, current, temperature, and charging capacity data generated by RCVAE trained with data from different cycles: (a) using data from the first 20 cycles; (b) using data from the first 40 cycles; (c) using data from the first 60 cycles; (d) using data from the first 80 cycles; (e) using data from the first 100 cycles. }
\label{fig1}
\end{figure}

To further explore the generative capabilities of RCVAE, Figure~\ref{fig1} displays electrochemical data samples generated by the RCVAE model trained using different amounts of early-cycle data, with these samples randomly selected from the corresponding test set. By specifying certain conditions (EOL+ECL), the model is capable of generating corresponding electrochemical data. For instance, Figure~\ref{fig1}a demonstrates the high consistency between the voltage, temperature, and charging capacity data generated by the RCVAE trained on the first 20 cycles and the original data, with $\text{R}^2$ values reaching at least 0.9715. For the current rate, although the quality of the generated data slightly decreased compared to other data types, the $\text{R}^2$ value was still 0.8771; Figure~\ref{fig1}b shows the data generated by the RCVAE trained on the first 40 cycles closely matching the original data in terms of voltage, current rate, temperature, and charging capacity, with $\text{R}^2$ values of at least 0.9616; In Figure~\ref{fig1}c, data generated by the RCVAE trained with the first 60 cycles accurately reflects the original data, with $\text{R}^2$ values of at least 0.9669; As shown in Figure~\ref{fig1}d, the voltage, temperature, and charging capacity data generated by the RCVAE trained with the first 80 cycles closely align with the original data, with $\text{R}^2$ values of at least 0.9352. However, for the current rate, compared to other data types, the quality of the generated data was lower, with an $\text{R}^2$ value of 0.7844; Figure~\ref{fig1}e illustrates the data generated by the RCVAE trained with the first 100 cycles almost perfectly matching the original data in terms of voltage, current rate, temperature, and charging capacity, with $\text{R}^2$ values of at least 0.9632. Overall, the RCVAE demonstrated good performance in generating various electrochemical data. Compared to traditional regression and classification algorithms, RCVAE can provide more comprehensive electrochemical information. Nonetheless, there remains room for improvement in the quality of generated data for the current rate.

\begin{figure}[H]
\centering
\includegraphics[width=\textwidth]{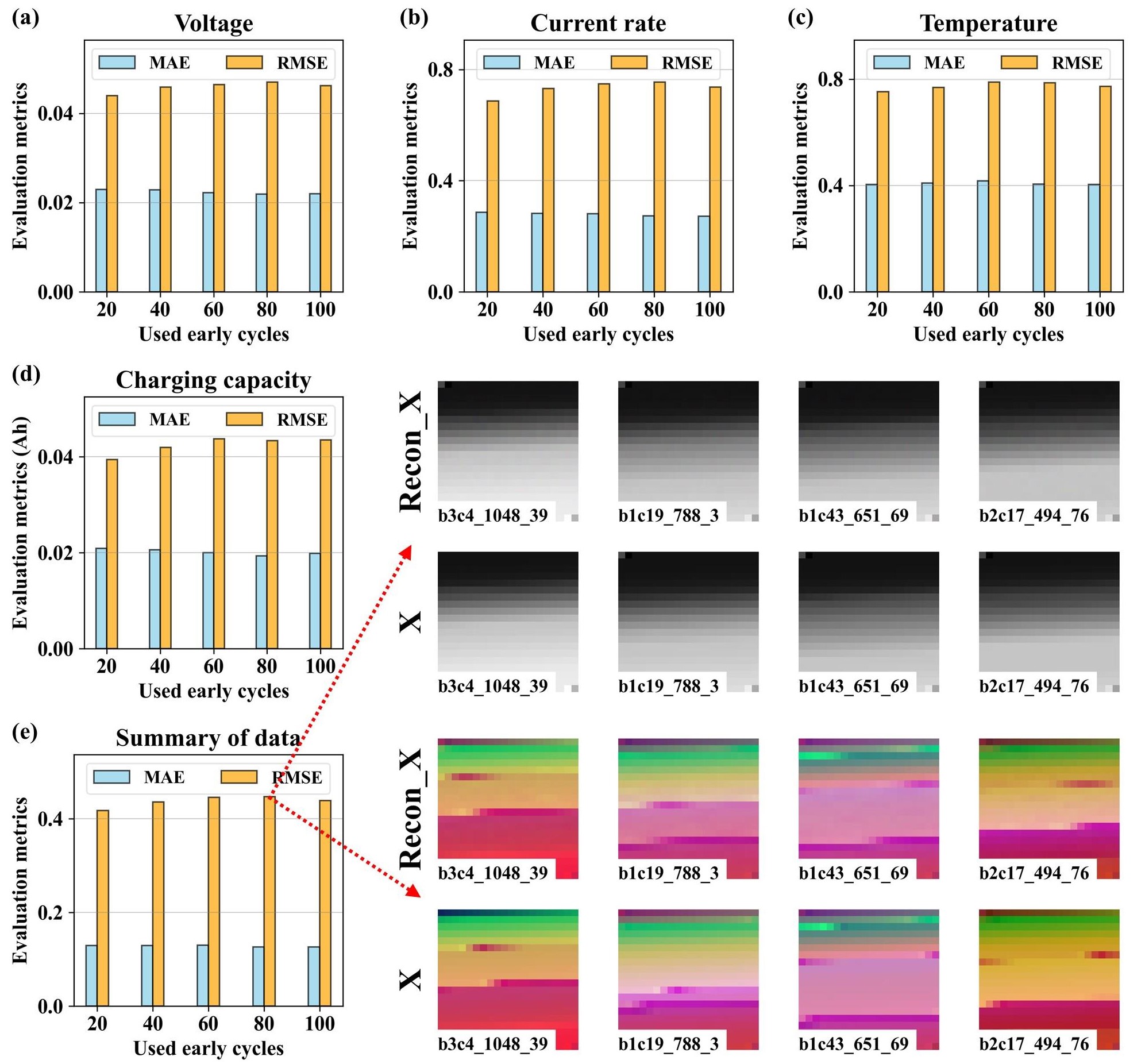} 
\caption{shows the statistical distribution of errors for different types of electrochemical data generated by RCVAE under various early-cycle conditions: (a) voltage error distribution, (b) current rate error distribution, (c) temperature error distribution, (d) charging capacity error distribution, and (e) a comprehensive summary of errors for all types of electrochemical data. }
\label{fig2}
\end{figure}

\unskip
To delve deeper into the variations in the generative capability of the RCVAE model with different amounts of early-cycle data, this study conducted a detailed visualization of error data, as shown in Figure~\ref{fig2}. In Figure~\ref{fig2}a, the capability of the RCVAE to generate voltage data across varying early-cycle data amounts is displayed, where the MAE slightly exceeds 0.02V, and the RMSE slightly exceeds 0.04V. Figure~\ref{fig2}b presents the RCVAE's ability to generate current rate data across different early-cycle data volumes, with the MAE values consistently less than 0.3 and the RMSE slightly over 0.6. In Figure~\ref{fig2}c, we observe the RCVAE's capacity to generate temperature data with varying early-cycle data amounts, where the MAE values remain under 0.3℃, and the RMSE slightly exceeds 0.6℃. Finally, Figure~\ref{fig2}d shows the RCVAE's ability to generate charging capacity data across different early-cycle data amounts, with MAE values fluctuating around 0.02Ah, and RMSE values around 0.04Ah.

Figure~\ref{fig2}e displays a summary of errors for various types of electrochemical charging data generated by the RCVAE across different amounts of early-cycle data. In all cases, the MAE remains below 0.2, while the RMSE slightly exceeds 0.4. Overall, the variation in MAE across different early-cycle data amounts is minimal, with RMSE reaching its peak when the early-cycle count is at 80. In this scenario, we further demonstrate the generative capability of the trained RCVAE model. By randomly selecting a condition (i.e., label) from the test set, the corresponding electrochemical charging data was successfully generated. For illustrative purposes and to present more generated samples, charging capacity data is scaled and presented in grayscale images. Voltage, current rate, and temperature data are combined and similarly scaled to be displayed in the form of RGB images. Taking the grayscale images as an example, the first row displays the generated charging capacity data, while the second row shows the original charging capacity data, with the battery number, EOL, and ECL labeled at the bottom left of each subplot. The results indicate that RCVAE can accurately generate corresponding high-quality charging capacity data for any specified EOL and ECL. The RGB images composed of voltage, current, and temperature also show similar compatibility, which is not elaborated here due to brevity. The number of generated samples shown is limited. To more comprehensively demonstrate the generative capabilities of RCVAE, Figure~\ref{figs1} and Figure~\ref{figs2} respectively present more sample details generated by RCVAE trained with different amounts of early-cycle data.

\subsection{Ablation Experiments of RCVAE}

\begin{figure}[H]
\centering
\includegraphics[width=\textwidth]{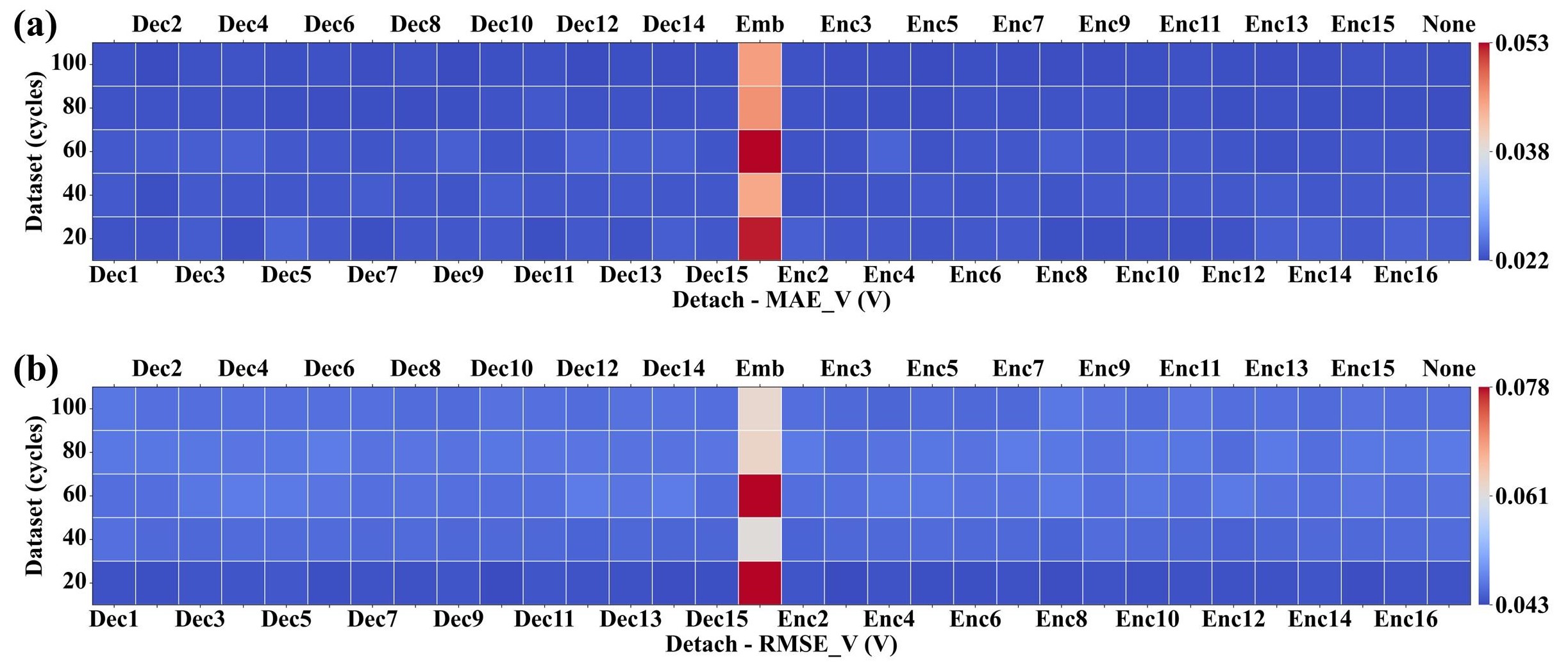} 
\caption{reveals the results of ablation experiments on RCVAE in generating voltage data.}
\label{fig3}
\end{figure}

To gain a deeper understanding of the contribution of each component within RCVAE, a series of ablation experiments were conducted. Figure~\ref{fig3}a illustrates the changes in error when generating voltage data after removing different layers of the model. Here, "Dec", "Enc", "Emb", and "None" respectively represent scenarios where the decoder, encoder, embedding layer were removed, and the standard RCVAE model without any modifications. Additionally, numbers following "Dec" and "Enc" indicate the specific number of layers removed. Given the critical role of the first layer of the encoder and the last layer of the decoder, experimental results reveal that even after removing any fully connected layer besides these two, the model's performance in generating voltage data remains excellent, demonstrating its high degree of robustness due to the numerous layers. Notably, the removal of the embedding layer significantly increased the MAE in generated data, with the MAE at least doubling (0.044-0.046V) under all early data training conditions; when trained with the first 20 and first 60 cycles of data, the MAE even rose to 0.053V. Figure~\ref{fig3}b shows the RMSE in generating voltage data, with trends similar to MAE, further confirming the model's architecture consisting of 16 fully connected layers in both the encoder and decoder. Despite the removal of a fully connected layer, the model's generative performance still remained good, showcasing its excellent robustness. However, the importance of the embedding layer is evident, as its removal significantly increased the RMSE in generated data, especially under the training conditions of the first 40, 80, and 100 cycles, where the RMSE (0.046-0.063V) nearly doubled. Particularly, under the training conditions of the first 20 and first 60 cycles, the MAE increased to 0.078V. Additionally, Figure~\ref{figs3}-Figure~\ref{figs5} respectively present heatmaps of MAE and RMSE in the ablation experiments for current rate, temperature data, and charging capacity data.

Moreover, Table~\ref{tab:tab2} meticulously details the errors in voltage and current rate data generated by the RCVAE trained with the first 100 cycles of data, along with the weighted errors for all data types, consistent with previous observations. The ablation experiments for voltage and current rate (Table~\ref{tabS1}), as well as temperature and charging capacity (Table~\ref{tabS2}), provide comprehensive ablation experiment data across different volumes of early-cycle data, once again highlighting the critical role of the embedding layer. These tables thoroughly confirm that the embedding layer is indispensable for maintaining the model's capability to produce high-quality data across the tasks of generating voltage, current rate, temperature, and charging capacity data.

\begin{table}[H]
\caption{Ablation Experiment Results of RCVAE Trained with Data from the First 100 Cycles }
\smallskip
\label{tab:tab2}
\begin{tabularx}{\textwidth}{YYYYYYY}
\toprule
\textbf{Detach} & \textbf{MAE} & \textbf{RMSE} & \textbf{MAE\_V (V)} & \textbf{RMSE\_V (V)} & \textbf{MAE\_I} & \textbf{RMSE\_I} \\
\midrule
Decoder\_1 & 0.127 & 0.447 & 0.022 & 0.047 & 0.273 & 0.749 \\
Decoder\_2 & 0.124 & 0.442 & 0.022 & 0.046 & 0.267 & 0.74 \\
Encoder\_2 & 0.126 & 0.437 & 0.022 & 0.046 & 0.272 & 0.735 \\
Decoder\_3 & 0.128 & 0.442 & 0.022 & 0.046 & 0.277 & 0.737 \\
Encoder\_3 & 0.127 & 0.437 & 0.022 & 0.046 & 0.274 & 0.726 \\
Decoder\_4 & 0.127 & 0.44 & 0.022 & 0.046 & 0.277 & 0.74 \\
Encoder\_4 & 0.125 & 0.434 & 0.022 & 0.045 & 0.273 & 0.729 \\
Decoder\_5 & 0.126 & 0.438 & 0.022 & 0.046 & 0.274 & 0.737 \\
Encoder\_5 & 0.125 & 0.438 & 0.022 & 0.046 & 0.271 & 0.731 \\
Decoder\_6 & 0.128 & 0.442 & 0.022 & 0.046 & 0.276 & 0.733 \\
Encoder\_6 & 0.126 & 0.436 & 0.022 & 0.046 & 0.275 & 0.733 \\
Decoder\_7 & 0.125 & 0.441 & 0.022 & 0.046 & 0.271 & 0.737 \\
Encoder\_7 & 0.127 & 0.437 & 0.022 & 0.046 & 0.274 & 0.728 \\
Decoder\_8 & 0.126 & 0.437 & 0.022 & 0.046 & 0.274 & 0.735 \\
Encoder\_8 & 0.125 & 0.447 & 0.022 & 0.047 & 0.265 & 0.747 \\
Decoder\_9 & 0.125 & 0.436 & 0.022 & 0.046 & 0.271 & 0.734 \\
Encoder\_9 & 0.126 & 0.445 & 0.022 & 0.047 & 0.271 & 0.747 \\
Decoder\_10 & 0.127 & 0.442 & 0.022 & 0.046 & 0.274 & 0.739 \\
Encoder\_10 & 0.127 & 0.438 & 0.022 & 0.046 & 0.276 & 0.732 \\
Decoder\_11 & 0.127 & 0.44 & 0.022 & 0.046 & 0.275 & 0.738 \\
Encoder\_11 & 0.127 & 0.444 & 0.022 & 0.047 & 0.273 & 0.747 \\
Decoder\_12 & 0.125 & 0.438 & 0.022 & 0.046 & 0.27 & 0.738 \\
Encoder\_12 & 0.126 & 0.441 & 0.022 & 0.046 & 0.273 & 0.738 \\
Decoder\_13 & 0.126 & 0.443 & 0.022 & 0.046 & 0.272 & 0.742 \\
Encoder\_13 & 0.125 & 0.44 & 0.022 & 0.046 & 0.27 & 0.739 \\
Decoder\_14 & 0.126 & 0.443 & 0.022 & 0.046 & 0.274 & 0.744 \\
Encoder\_14 & 0.126 & 0.436 & 0.022 & 0.046 & 0.272 & 0.735 \\
Decoder\_15 & 0.126 & 0.438 & 0.022 & 0.046 & 0.272 & 0.738 \\
Encoder\_15 & 0.127 & 0.44 & 0.022 & 0.046 & 0.274 & 0.737 \\
Encoder\_16 & 0.128 & 0.44 & 0.022 & 0.046 & 0.273 & 0.732 \\
None & 0.126 & 0.439 & 0.022 & 0.046 & 0.272 & 0.737 \\
Embedding & 0.296 & 0.635 & 0.045 & 0.062 & 0.652 & 0.955 \\
\bottomrule
\end{tabularx}
\end{table}

\subsection{Learning Outcomes of the Embedding Layer}

\begin{figure}[H]
\centering
\includegraphics[width=\textwidth]{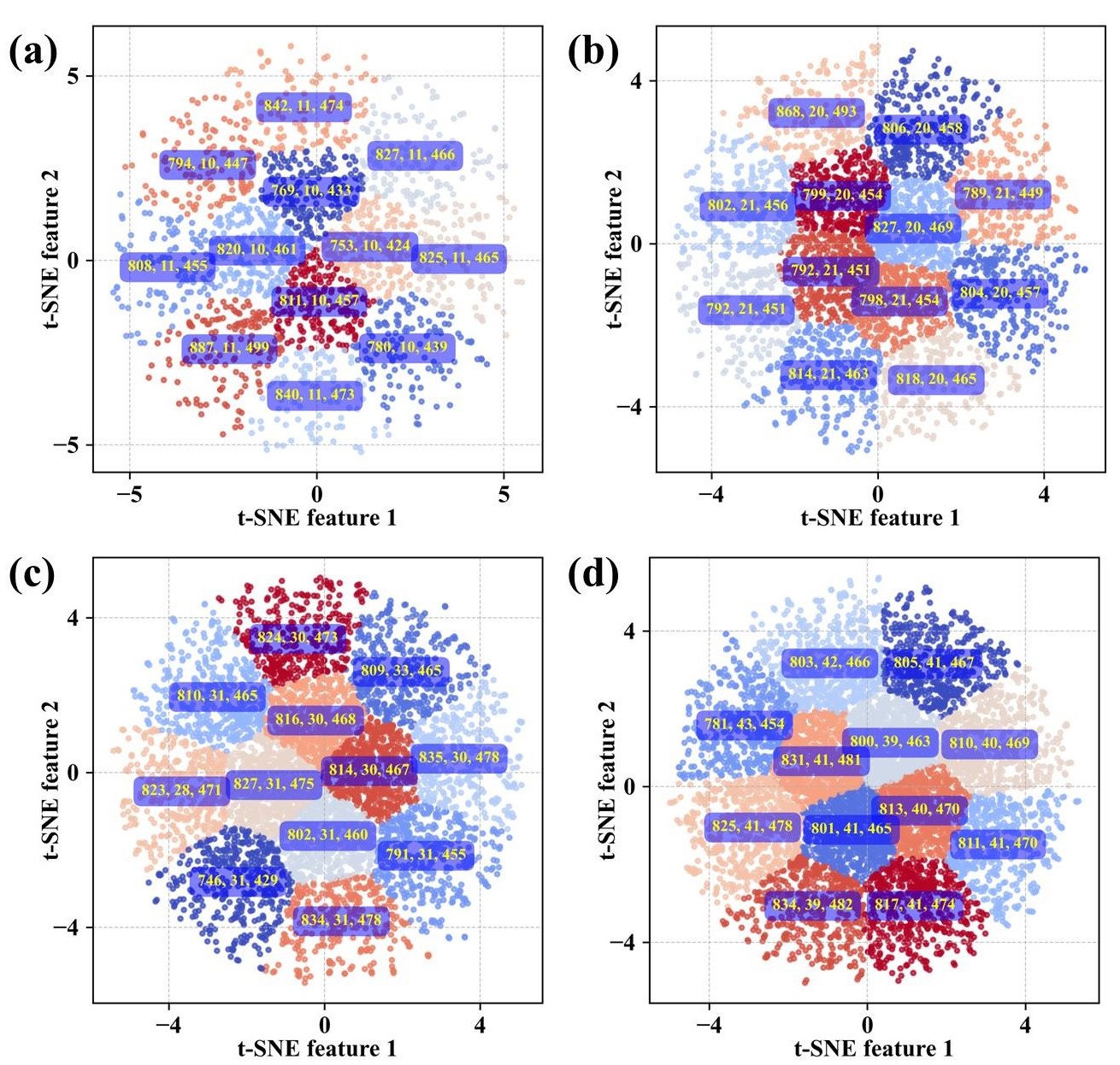}
\caption{ presents the learning outcomes of the RCVAE embedding layer trained with data from different cycles: (a) using data from the first 20 cycles; (b) using data from the first 40 cycles; (c) using data from the first 60 cycles; (d) using data from the first 80 cycles. }
\label{fig4}
\end{figure}

Following the discussion on the importance of the embedding layer for RCVAE in generating electrochemical data in the previous section, this section delves further into the mechanism of the embedding layer's function and elucidates it through the visualization of learned results. Figure~\ref{fig4} displays the distribution of weights learned in the embedding layer. Given the multitude of conditional categories caused by the combination of EOL and ECL, the dimensionality of the embedding vectors is correspondingly high (embedding\_dim=473). To simplify the presentation, we first utilized t-SNE\cite{maaten2008visualizing} to reduce the dimensionality of the embedding vectors to a two-dimensional space, and then performed clustering analysis using the KNN algorithm\cite{cover1967nearest}.

Figure~\ref{fig4}a presents the clustering results of the model's embedding vectors trained on the first 20 cycles of data. Each cluster is annotated with three values: the average EOL within the cluster, the average ECL, and the weighted value calculated using equation (\ref{eq14}), which represents the distance from the condition where both EOL and ECL are zero. Considering the model was trained on data from the first 20 cycles, the ECL represented within the clusters are generally around 10 or 11, aligning with the actual variation patterns of electrochemical data in batteries: samples with different EOLs exhibit significant electrochemical data differences. If the EOL is similar and the ECL is close, the variation in the battery's electrochemical data is smaller, showing a gradual change, as demonstrated and verified in Figure~\ref{fig6}a. In other words, under the conditions formed by both EOL and ECL in the sample generation process, EOL plays a leading role, while ECL acts as an auxiliary modulator.

Figure~\ref{fig4}b displays the distribution of the model's embedding vectors trained on the first 40 cycles of data. Compared to the training with only the first 20 cycles of data, the number of labels has doubled, leading to a significant change in the distribution of embedding vectors. The range of the average EOL values for different clusters (79=868-789) has become more compact compared to the range during the first 20 cycles of data (134=887-753). Meanwhile, ECL differences within clusters (from 1=11-10 to 1=21-20) have not changed significantly. This suggests that the dominant effect of EOL has diminished, thereby relatively enhancing the auxiliary role of ECL.

Figure~\ref{fig4}c illustrates the distribution of the model's embedding vectors trained with the first 60 cycles of data. Compared to the training that utilized only the first 20 cycles of data, the number of labels has tripled, leading to a noticeable change in the distribution of embedding vectors. The range of average EOL values across different clusters (89=835-746) has become more compact compared to the range during the first 20 cycles (134=887-753). Additionally, ECL differences within clusters have become more pronounced, expanding from 1 (11-10) to 5 (33-28). Considering the current range of ECL is from 60 to 1, as opposed to the previous 20 to 1, the variation in electrochemical data corresponding to ECL has increased, which results in the relative weakening of EOL's dominant effect and an enhancement of ECL's auxiliary role. Comparing Figure~\ref{fig4}b and Figure~\ref{fig4}c, the average EOL range shifted from (79=868-789) to (89=835-746), and the differences in cycle numbers within clusters expanded from 1 (21-20) to 5 (33-28). The differences between EOL values have not changed significantly, but the differences between ECL values have quadrupled, indicating an enhancement in the auxiliary role of ECL at this stage.

Figure~\ref{fig4}d displays the distribution of conditions for the RCVAE when utilizing data from the first 80 cycles, where the range of EOL values within different clusters (53=834-781) has significantly decreased compared to earlier stages. The range of ECL (4=43-39) has not changed much from the scenario with the first 60 cycles but shows significant variation from the first 20 and 40 cycles. This pattern, consistent with the trends observed in Figure~\ref{fig4}a-c, demonstrates that as the number of early cycles used increases, the influence of EOL in directing data generation under supervision gradually diminishes, while the role of ECL in the conditions becomes increasingly significant. Figure~\ref{figs6} then shows the distribution of embedding layer weights for the RCVAE trained with data from the first 100 cycles, maintaining consistency with the patterns observed in Figure~\ref{fig4}. These findings indicate that under different data distribution scenarios, the combined conditions of EOL and ECL work together effectively to adapt to different scenario needs, supervising the generation of electrochemical data effectively. This further highlights the importance of the embedding layer and its central role within the RCVAE model.

\section{Discussion}

By integrating the embedding layer into the CVAE model, this study successfully developed the RCVAE, aimed specifically at enhancing the generative capability for electrochemical data. The research adopted a quasi-video data preprocessing method, effectively integrating different types of electrochemical data (such as voltage, current, temperature, and charging capacity), to serve as input for the RCVAE. In this work, EOL and ECL were defined as conditions for the generative artificial intelligence, enabling the generative model to target and supervise the production of electrochemical data corresponding to specific EOLs, thereby incorporating information about remaining lifespan.

Experimental results confirm that RCVAE can accurately generate a variety of charging data. Under conditions trained with different amounts of early-cycle data, RCVAE consistently demonstrated exceptional capabilities in generating electrochemical data. Taking voltage data as an example, when utilizing the first 20, 40, 60, 80, and 100 cycles of data as the training set, the MAE reached 0.023V, 0.023V, 0.022V, 0.022V, and 0.022V respectively; the RMSE was 0.044V, 0.046V, 0.046V, 0.047V, and 0.046V, respectively. Ablation experiments on RCVAE revealed the model's high robustness, with its performance remaining stable even after removing a layer. Notably, the significant reduction in generative performance after the removal of the embedding layer emphasizes the importance of this modification to the CVAE model.

Further analysis of the learning outcomes of the embedding layer reiterated the critical role of the embedding layer and the significant importance of combining EOL and ECL as conditions. Given the similarity in data characteristics observed in other tasks, this suggests that RCVAE has the potential for broad applications across a wider range of fields.

\section*{Declaration of Competing Interest}
The authors declare that they have no known competing financial interests or personal relationships that could have appeared to influence the work reported in this paper.

\section*{Data Availability}
Data will be made available on request.

\section*{Nomenclature}
\begin{tabular}{@{}ll}
EOL & End of Life \\
ECL & Equivalent Cycle Life \\
RCVAE & Refined Conditional Variational Autoencoder \\
LIBs & Lithium-ion batteries \\
RNN & Recurrent Neural Networks \\
CNN & Convolutional Neural Networks \\
NLP & natural language processing \\
CV & computer vision \\
GANs & Generative Adversarial Networks \\
ViT & Vision Transformer \\
VAE & Variational Autoencoders \\
CVAE & Conditional Variational Autoencoders \\
MAPE & Mean Absolute Percentage Error \\
MAE & Mean Absolute Error \\
RMSE & Root Mean Square Error \\
BMS & Battery Management System \\
RUL & Remaining Useful Lives \\
MLP & multi-layer fully connected neural network \\
\end{tabular}

\bibliography{references}
\clearpage 

\appendix
\setcounter{figure}{0}
\renewcommand{\thefigure}{S\arabic{figure}}
\setcounter{table}{0}
\renewcommand{\thetable}{S\arabic{table}}

\section*{\centering Supporting Information}
\begin{center}
\textbf{Generating Comprehensive Lithium Battery Charging Data with Generative AI}

Lidang Jiang${}^{\ a}$, Changyan Hu${}^{\ a}$, Sibei Ji${}^{\ a}$, Hang Zhao${}^{\ a}$, Junxiong Chen${}^{\ b}$*, Ge He${}^{\ a}$*

${}^{a}$ School of Chemical Engineering, Sichuan University, Chengdu 610000 Sichuan, PR China

${}^{b}$ School of Automation Engineering, University of Electronic Science and Technology of China, Chengdu, 610023, China
\end{center}

\section*{Supplementary Notes on Figures}
\vspace{-25pt}
\begin{figure}[H]
\centering
\includegraphics[width=\textwidth]{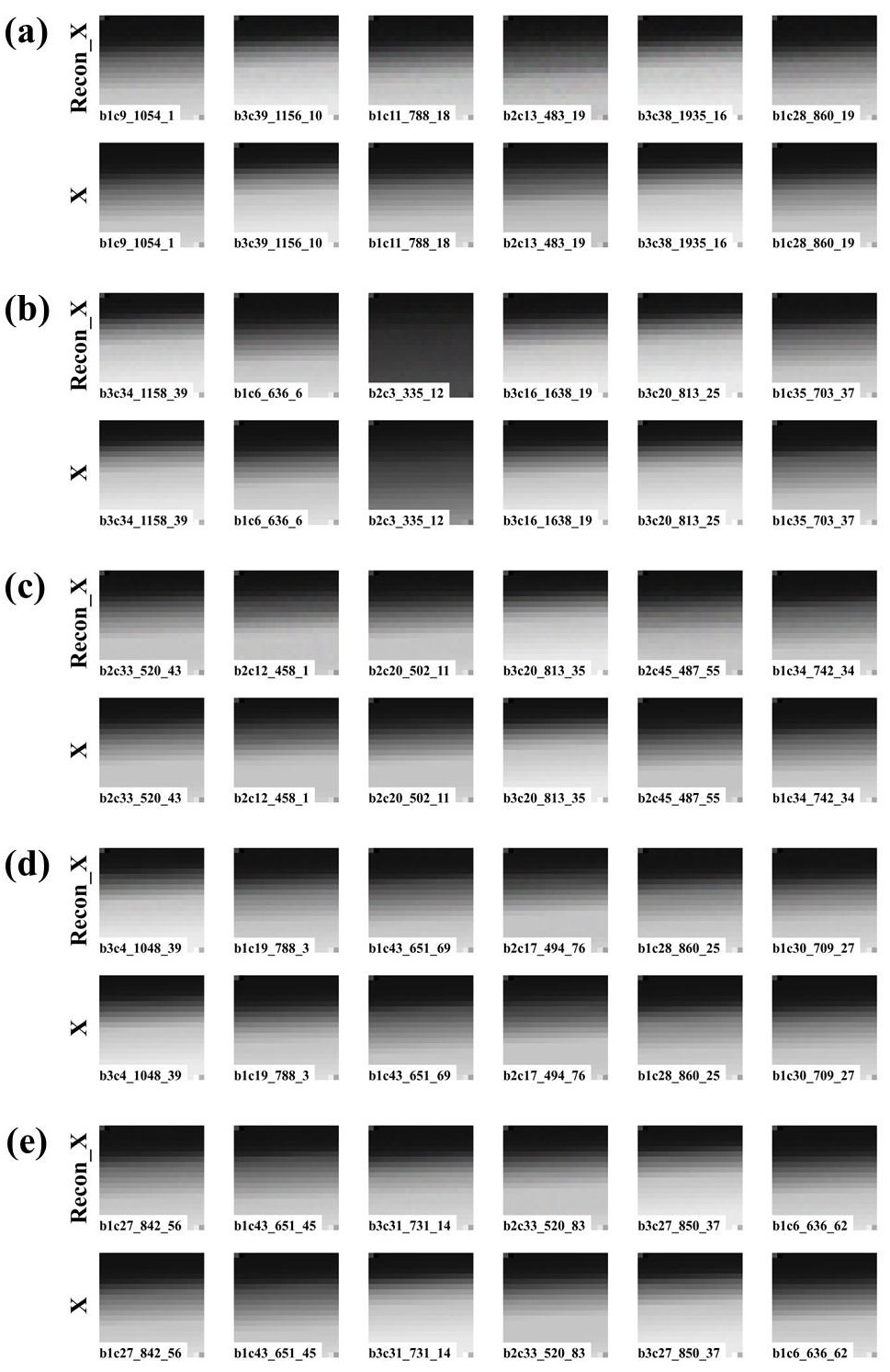}
\caption{shows grayscale images of scaled charging capacity generated by RCVAE trained with data from different cycles: (a) using data from the first 20 cycles; (b) using data from the first 40 cycles; (c) using data from the first 60 cycles; (d) using data from the first 80 cycles; (e) using data from the first 100 cycles. }
\label{figs1}
\end{figure}

\begin{figure}[H]
\centering
\includegraphics[width=\textwidth]{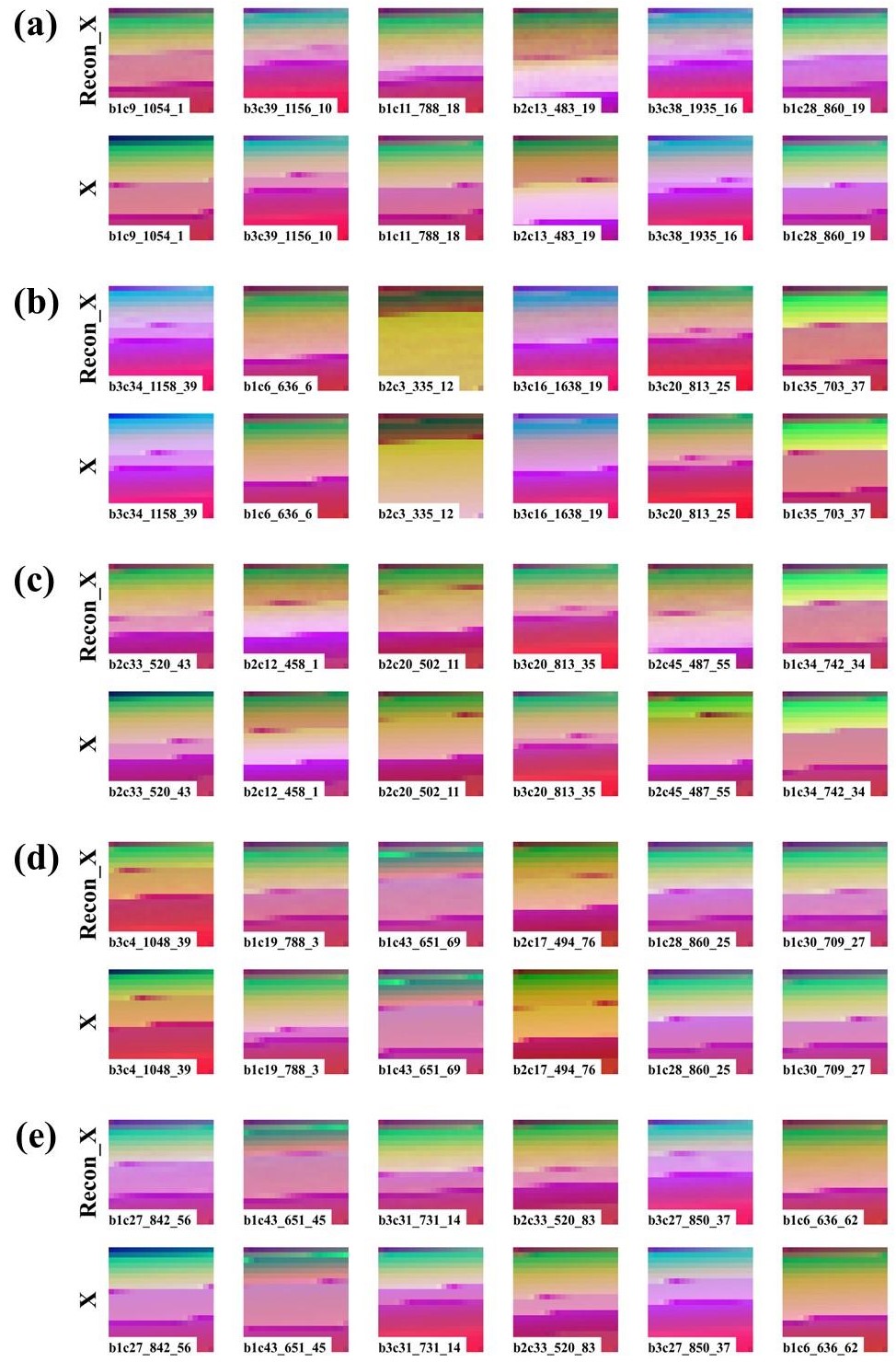}
\caption{presents RGB images synthesized from voltage, current, and temperature data generated by RCVAE trained with data from different cycles: (a) using data from the first 20 cycles; (b) using data from the first 40 cycles; (c) using data from the first 60 cycles; (d) using data from the first 80 cycles; (e) using data from the first 100 cycles.}
\label{figs2}
\end{figure}

\begin{figure}[H]
\centering
\includegraphics[width=\textwidth]{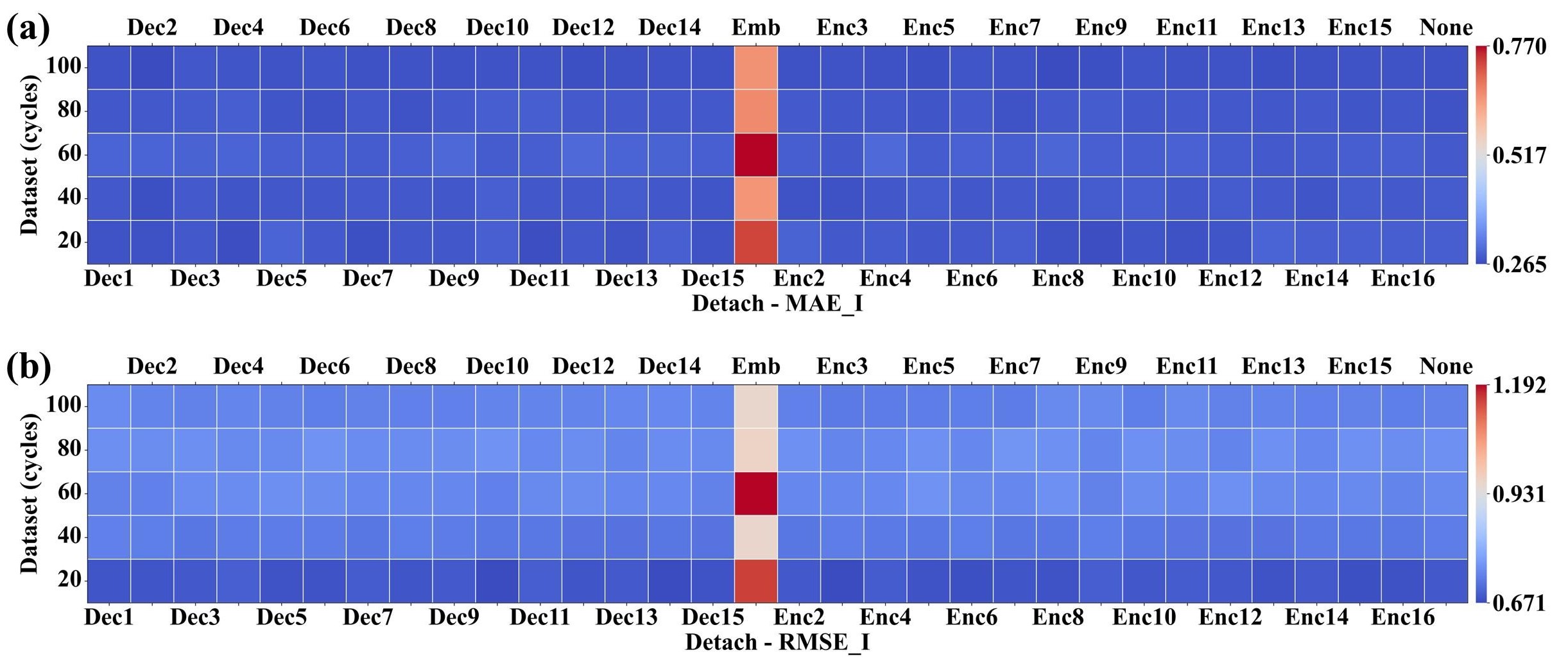}
\caption{shows the results of ablation experiments on RCVAE in generating current rate data. }
\label{figs3}
\end{figure}

\begin{figure}[H]
\centering
\includegraphics[width=\textwidth]{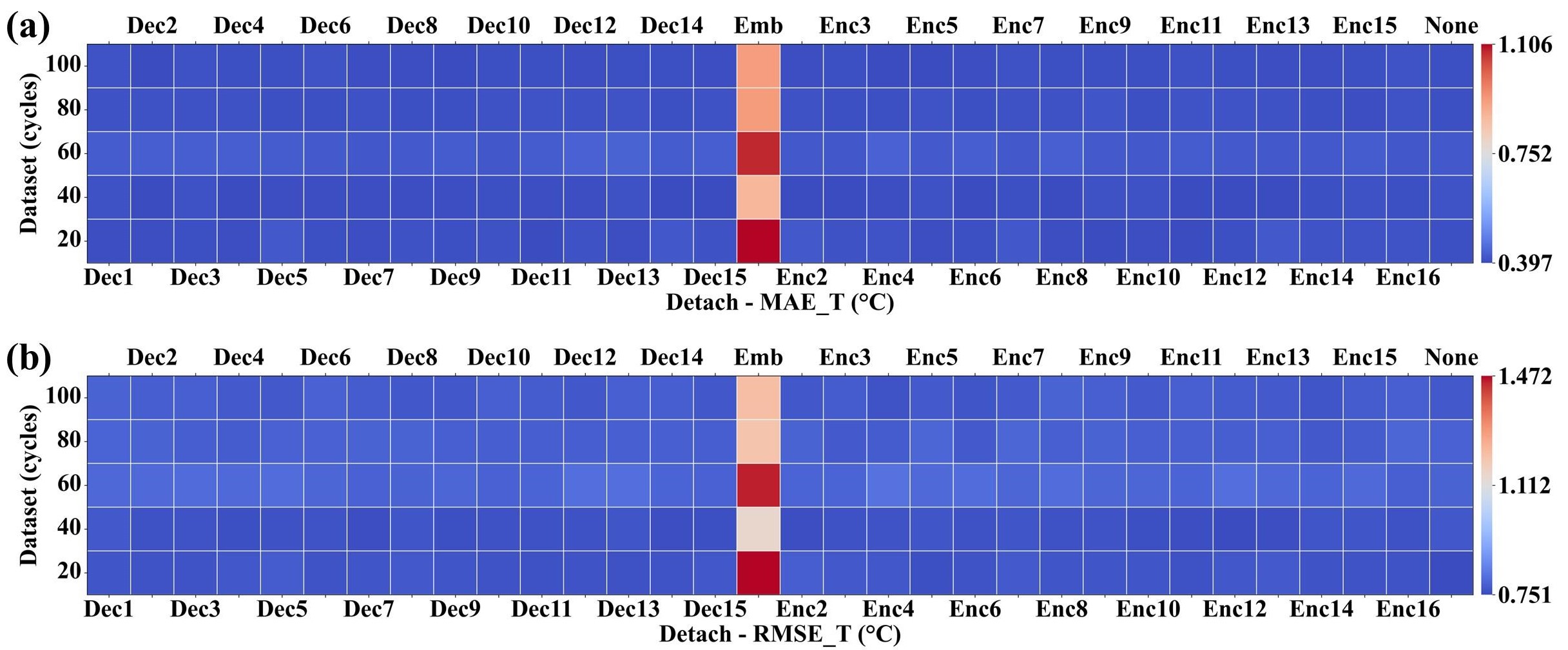}
\caption{presents the results of ablation experiments on RCVAE in generating temperature data. }
\label{figs4}
\end{figure}

\begin{figure}[H]
\centering
\includegraphics[width=\textwidth]{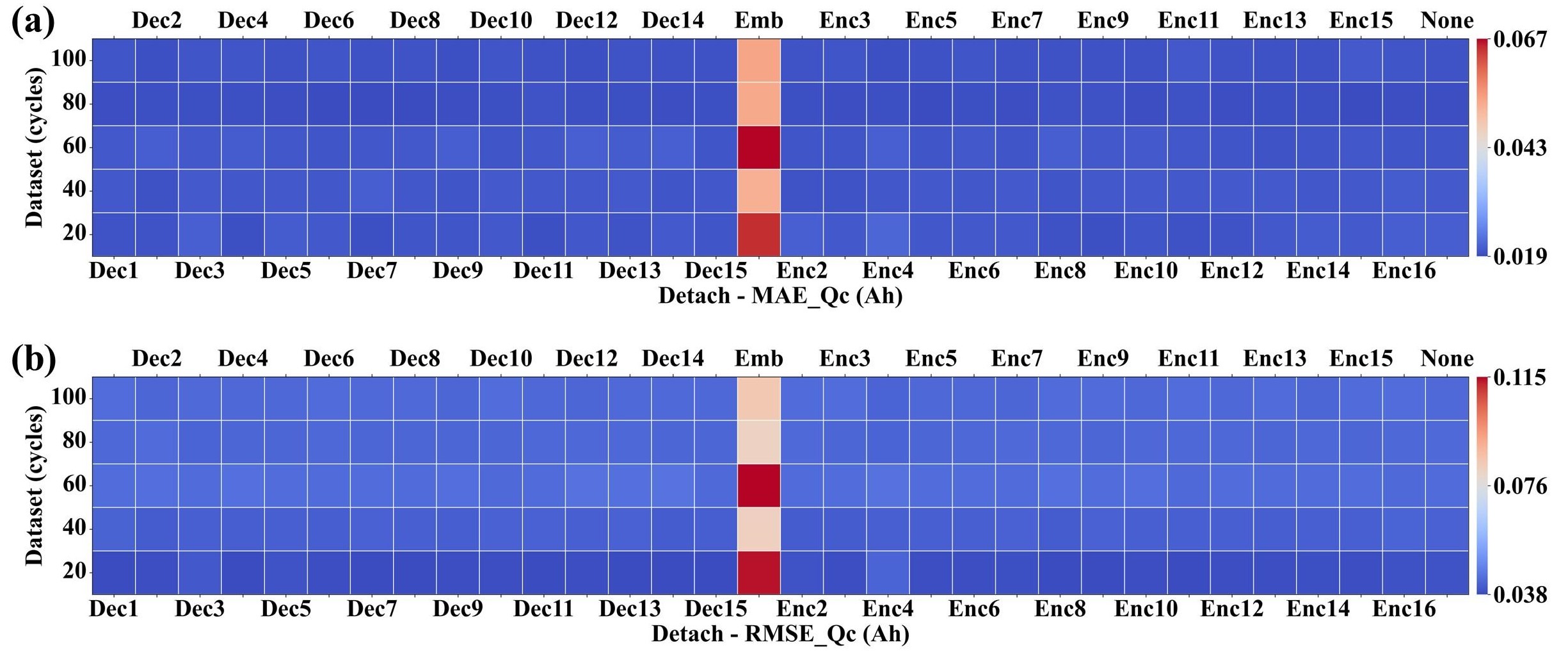}
\caption{displays the results of ablation experiments on RCVAE in generating charging capacity data. }
\label{figs5}
\end{figure}

\begin{figure}[H]
\centering
\includegraphics[width=0.5\textwidth]{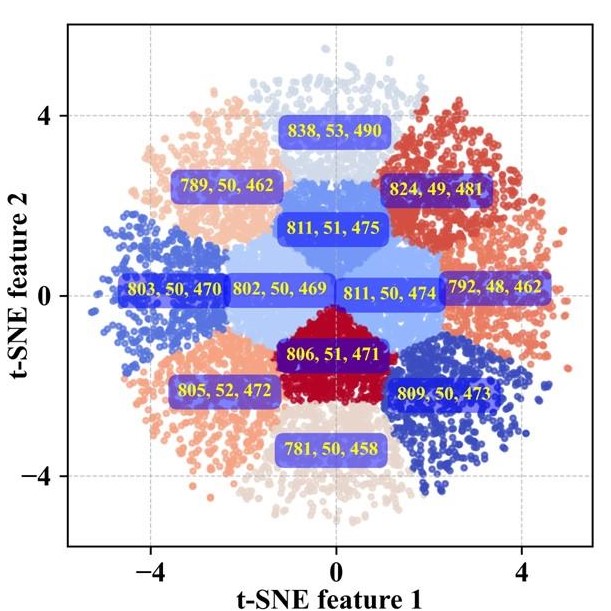} 
\caption{reveals the learning outcomes of the RCVAE embedding layer trained with data from the first 100 cycles. }
\label{figs6}
\end{figure}
\unskip
\section*{Supplementary Notes on Tables}
\vspace{-30pt}
\begin{table}[H]
\caption{Ablation Experiment Results for Voltage and Current Rate}\label{tabS1}
\smallskip

\begin{tabularx}{\textwidth}{YYYYYY}
\toprule
\textbf{Used Early Cycles} & \textbf{Detach} & \textbf{MAE\_V (V)} & \textbf{RMSE\_V (V)} & \textbf{MAE\_I} & \textbf{RMSE\_I} \\
\midrule
\multirow{32}{*}{20} & Decoder\_1 & 0.022 & 0.044 & 0.273 & 0.682 \\
& Decoder\_2 & 0.022 & 0.044 & 0.272 & 0.682 \\  
& Encoder\_2 & 0.023 & 0.044 & 0.292 & 0.686 \\  
& Decoder\_3 & 0.023 & 0.044 & 0.28 & 0.687 \\  
& Encoder\_3 & 0.022 & 0.043 & 0.28 & 0.672 \\  
& Decoder\_4 & 0.022 & 0.044 & 0.268 & 0.695 \\  
& Encoder\_4 & 0.023 & 0.044 & 0.284 & 0.69 \\  
& Decoder\_5 & 0.023 & 0.044 & 0.294 & 0.679 \\  
& Encoder\_5 & 0.022 & 0.044 & 0.278 & 0.68 \\  
& Decoder\_6 & 0.023 & 0.044 & 0.283 & 0.68 \\  
& Encoder\_6 & 0.023 & 0.044 & 0.281 & 0.676 \\  
& Decoder\_7 & 0.022 & 0.044 & 0.27 & 0.691 \\  
& Encoder\_7 & 0.023 & 0.044 & 0.285 & 0.682 \\  
& Decoder\_8 & 0.022 & 0.044 & 0.278 & 0.683 \\  
& Encoder\_8 & 0.022 & 0.043 & 0.272 & 0.68 \\  
& Decoder\_9 & 0.022 & 0.044 & 0.278 & 0.689 \\  
& Encoder\_9 & 0.022 & 0.044 & 0.269 & 0.692 \\  
& Decoder\_10 & 0.023 & 0.043 & 0.287 & 0.671 \\  
& Encoder\_10 & 0.022 & 0.044 & 0.275 & 0.684 \\  
& Decoder\_11 & 0.022 & 0.044 & 0.268 & 0.691 \\  
& Encoder\_11 & 0.022 & 0.044 & 0.271 & 0.691 \\  
& Decoder\_12 & 0.023 & 0.044 & 0.28 & 0.682 \\  
& Encoder\_12 & 0.022 & 0.044 & 0.276 & 0.686 \\  
& Decoder\_13 & 0.022 & 0.044 & 0.274 & 0.687 \\  
& Encoder\_13 & 0.023 & 0.044 & 0.293 & 0.68 \\  
& Decoder\_14 & 0.023 & 0.043 & 0.288 & 0.672 \\  
& Encoder\_14 & 0.023 & 0.044 & 0.288 & 0.688 \\  
& Decoder\_15 & 0.022 & 0.044 & 0.274 & 0.679 \\  
& Encoder\_15 & 0.023 & 0.043 & 0.287 & 0.675 \\  
& Encoder\_16 & 0.023 & 0.044 & 0.286 & 0.679 \\  
& None & 0.023 & 0.044 & 0.286 & 0.686 \\  
& Embedding & 0.053 & 0.078 & 0.732 & 1.157 \\  
\bottomrule
\end{tabularx}
\end{table}

\begin{table}[H]
\ContinuedFloat 
\captionsetup{labelsep=period}
\caption{{\em Cont.}}
\smallskip
\begin{tabularx}{\textwidth}{YYYYYY}
\toprule
\textbf{Used Early Cycles} & \textbf{Detach} & \textbf{MAE\_V (V)} & \textbf{RMSE\_V (V)} & \textbf{MAE\_I} & \textbf{RMSE\_I} \\
\midrule
\multirow{32}{*}{40} & Decoder\_1 & 0.023 & 0.046 & 0.281 & 0.735 \\  
& Decoder\_2 & 0.022 & 0.046 & 0.27 & 0.734 \\  
& Encoder\_2 & 0.022 & 0.045 & 0.273 & 0.723 \\  
& Decoder\_3 & 0.023 & 0.046 & 0.281 & 0.722 \\  
& Encoder\_3 & 0.022 & 0.046 & 0.273 & 0.731 \\  
& Decoder\_4 & 0.022 & 0.046 & 0.276 & 0.729 \\  
& Encoder\_4 & 0.022 & 0.046 & 0.277 & 0.727 \\  
& Decoder\_5 & 0.022 & 0.046 & 0.278 & 0.728 \\  
& Encoder\_5 & 0.023 & 0.046 & 0.284 & 0.725 \\  
& Decoder\_6 & 0.022 & 0.046 & 0.277 & 0.73 \\  
& Encoder\_6 & 0.022 & 0.046 & 0.278 & 0.732 \\  
& Decoder\_7 & 0.023 & 0.046 & 0.281 & 0.723 \\  
& Encoder\_7 & 0.023 & 0.046 & 0.282 & 0.722 \\  
& Decoder\_8 & 0.022 & 0.046 & 0.279 & 0.733 \\  
& Encoder\_8 & 0.022 & 0.045 & 0.277 & 0.722 \\  
& Decoder\_9 & 0.022 & 0.046 & 0.277 & 0.73 \\  
& Encoder\_9 & 0.023 & 0.046 & 0.283 & 0.733 \\  
& Decoder\_10 & 0.023 & 0.046 & 0.287 & 0.724 \\  
& Encoder\_10 & 0.023 & 0.046 & 0.285 & 0.724 \\  
& Decoder\_11 & 0.023 & 0.046 & 0.278 & 0.725 \\  
& Encoder\_11 & 0.023 & 0.045 & 0.279 & 0.72 \\  
& Decoder\_12 & 0.023 & 0.045 & 0.279 & 0.718 \\  
& Encoder\_12 & 0.023 & 0.045 & 0.284 & 0.717 \\  
& Decoder\_13 & 0.023 & 0.046 & 0.283 & 0.717 \\  
& Encoder\_13 & 0.023 & 0.045 & 0.282 & 0.719 \\  
& Decoder\_14 & 0.022 & 0.045 & 0.278 & 0.724 \\  
& Encoder\_14 & 0.022 & 0.046 & 0.276 & 0.728 \\  
& Decoder\_15 & 0.022 & 0.045 & 0.276 & 0.719 \\  
& Encoder\_15 & 0.023 & 0.046 & 0.284 & 0.724 \\  
& Encoder\_16 & 0.023 & 0.046 & 0.282 & 0.725 \\  
& None & 0.023 & 0.046 & 0.283 & 0.731 \\  
& Embedding & 0.044 & 0.061 & 0.649 & 0.957 \\  
\midrule
\multirow{21}{*}{60} & Decoder\_1 & 0.023 & 0.046 & 0.293 & 0.736 \\  
& Decoder\_2 & 0.023 & 0.046 & 0.293 & 0.735 \\  
& Encoder\_2 & 0.022 & 0.046 & 0.285 & 0.742 \\  
& Decoder\_3 & 0.023 & 0.047 & 0.291 & 0.749 \\  
& Encoder\_3 & 0.022 & 0.046 & 0.281 & 0.743 \\  
& Decoder\_4 & 0.023 & 0.047 & 0.294 & 0.749 \\  
& Encoder\_4 & 0.023 & 0.047 & 0.3 & 0.746 \\  
& Decoder\_5 & 0.023 & 0.047 & 0.287 & 0.753 \\  
& Encoder\_5 & 0.022 & 0.047 & 0.285 & 0.759 \\  
& Decoder\_6 & 0.022 & 0.047 & 0.287 & 0.752 \\  
& Encoder\_6 & 0.023 & 0.047 & 0.291 & 0.747 \\  
& Decoder\_7 & 0.022 & 0.046 & 0.285 & 0.743 \\  
& Encoder\_7 & 0.022 & 0.046 & 0.285 & 0.743 \\  
& Decoder\_8 & 0.023 & 0.046 & 0.289 & 0.743 \\  
& Encoder\_8 & 0.023 & 0.047 & 0.296 & 0.755 \\ 
& Decoder\_9 & 0.023 & 0.046 & 0.297 & 0.743 \\  
& Encoder\_9 & 0.023 & 0.046 & 0.289 & 0.737 \\  
& Decoder\_10 & 0.022 & 0.046 & 0.284 & 0.735 \\  
& Encoder\_10 & 0.023 & 0.047 & 0.285 & 0.751 \\  
& Decoder\_11 & 0.022 & 0.046 & 0.287 & 0.747 \\  
& Encoder\_11 & 0.023 & 0.046 & 0.29 & 0.743 \\  
\bottomrule
\end{tabularx}
\end{table}

\begin{table}[H]
\ContinuedFloat 
\captionsetup{labelsep=period}
\caption{{\em Cont.}}
\smallskip
\begin{tabularx}{\textwidth}{YYYYYY}
\toprule
\textbf{Used Early Cycles} & \textbf{Detach} & \textbf{MAE\_V (V)} & \textbf{RMSE\_V (V)} & \textbf{MAE\_I} & \textbf{RMSE\_I} \\
\midrule
\multirow{11}{*}{60} & Decoder\_12 & 0.023 & 0.047 & 0.299 & 0.751 \\  
& Encoder\_12 & 0.022 & 0.047 & 0.283 & 0.755 \\  
& Decoder\_13 & 0.023 & 0.047 & 0.294 & 0.743 \\  
& Encoder\_13 & 0.022 & 0.046 & 0.281 & 0.747 \\  
& Decoder\_14 & 0.023 & 0.047 & 0.292 & 0.746 \\  
& Encoder\_14 & 0.022 & 0.046 & 0.284 & 0.743 \\  
& Decoder\_15 & 0.022 & 0.046 & 0.287 & 0.737 \\  
& Encoder\_15 & 0.022 & 0.047 & 0.286 & 0.747 \\  
& Encoder\_16 & 0.022 & 0.046 & 0.287 & 0.739 \\  
& None & 0.022 & 0.046 & 0.282 & 0.748 \\  
& Embedding & 0.053 & 0.078 & 0.77 & 1.192 \\
\midrule
\multirow{32}{*}{80} & Decoder\_1 & 0.022 & 0.047 & 0.279 & 0.754 \\  
& Decoder\_2 & 0.022 & 0.047 & 0.281 & 0.751 \\  
& Encoder\_2 & 0.022 & 0.047 & 0.281 & 0.755 \\  
& Decoder\_3 & 0.022 & 0.047 & 0.283 & 0.754 \\  
& Encoder\_3 & 0.022 & 0.046 & 0.278 & 0.741 \\  
& Decoder\_4 & 0.022 & 0.047 & 0.285 & 0.747 \\  
& Encoder\_4 & 0.022 & 0.046 & 0.28 & 0.746 \\  
& Decoder\_5 & 0.022 & 0.047 & 0.275 & 0.747 \\  
& Encoder\_5 & 0.022 & 0.047 & 0.277 & 0.755 \\  
& Decoder\_6 & 0.022 & 0.047 & 0.272 & 0.757 \\  
& Encoder\_6 & 0.022 & 0.047 & 0.281 & 0.743 \\  
& Decoder\_7 & 0.022 & 0.047 & 0.28 & 0.749 \\  
& Encoder\_7 & 0.022 & 0.047 & 0.275 & 0.762 \\  
& Decoder\_8 & 0.022 & 0.047 & 0.274 & 0.749 \\  
& Encoder\_8 & 0.022 & 0.047 & 0.282 & 0.754 \\  
& Decoder\_9 & 0.022 & 0.046 & 0.282 & 0.751 \\  
& Encoder\_9 & 0.022 & 0.046 & 0.284 & 0.74 \\  
& Decoder\_10 & 0.022 & 0.047 & 0.286 & 0.757 \\  
& Encoder\_10 & 0.022 & 0.047 & 0.279 & 0.755 \\  
& Decoder\_11 & 0.023 & 0.047 & 0.286 & 0.745 \\  
& Encoder\_11 & 0.022 & 0.047 & 0.282 & 0.751 \\  
& Decoder\_12 & 0.022 & 0.047 & 0.282 & 0.751 \\  
& Encoder\_12 & 0.022 & 0.046 & 0.281 & 0.74 \\  
& Decoder\_13 & 0.022 & 0.046 & 0.283 & 0.741 \\  
& Encoder\_13 & 0.022 & 0.047 & 0.278 & 0.756 \\  
& Decoder\_14 & 0.022 & 0.047 & 0.28 & 0.749 \\  
& Encoder\_14 & 0.022 & 0.046 & 0.281 & 0.743 \\  
& Decoder\_15 & 0.022 & 0.047 & 0.28 & 0.747 \\  
& Encoder\_15 & 0.022 & 0.047 & 0.276 & 0.755 \\  
& Encoder\_16 & 0.022 & 0.047 & 0.278 & 0.75 \\  
& None & 0.022 & 0.047 & 0.273 & 0.755 \\  
& Embedding & 0.046 & 0.063 & 0.662 & 0.965 \\ 
\midrule
\multirow{8}{*}{100} & Decoder\_1 & 0.022 & 0.047 & 0.273 & 0.749 \\  
& Decoder\_2 & 0.022 & 0.046 & 0.267 & 0.74 \\  
& Encoder\_2 & 0.022 & 0.046 & 0.272 & 0.735 \\  
& Decoder\_3 & 0.022 & 0.046 & 0.277 & 0.737 \\  
& Encoder\_3 & 0.022 & 0.046 & 0.274 & 0.726 \\  
& Decoder\_4 & 0.022 & 0.046 & 0.277 & 0.74 \\ 
& Encoder\_4 & 0.022 & 0.045 & 0.273 & 0.729 \\  
& Decoder\_5 & 0.022 & 0.046 & 0.274 & 0.737 \\  
& Encoder\_5 & 0.022 & 0.046 & 0.271 & 0.731 \\ 
\bottomrule
\end{tabularx}
\end{table}

\begin{table}[H]
\ContinuedFloat 
\captionsetup{labelsep=period}
\caption{{\em Cont.}}
\smallskip
\begin{tabularx}{\textwidth}{YYYYYY}
\toprule
\textbf{Used Early Cycles} & \textbf{Detach} & \textbf{MAE\_V (V)} & \textbf{RMSE\_V (V)} & \textbf{MAE\_I} & \textbf{RMSE\_I} \\
\midrule
\multirow{23}{*}{100} & Decoder\_6 & 0.022 & 0.046 & 0.276 & 0.733 \\  
& Encoder\_6 & 0.022 & 0.046 & 0.275 & 0.733 \\  
& Decoder\_7 & 0.022 & 0.046 & 0.271 & 0.737 \\  
& Encoder\_7 & 0.022 & 0.046 & 0.274 & 0.728 \\  
& Decoder\_8 & 0.022 & 0.046 & 0.274 & 0.735 \\  
& Encoder\_8 & 0.022 & 0.047 & 0.265 & 0.747 \\  
& Decoder\_9 & 0.022 & 0.046 & 0.271 & 0.734 \\  
& Encoder\_9 & 0.022 & 0.047 & 0.271 & 0.747 \\  
& Decoder\_10 & 0.022 & 0.046 & 0.274 & 0.739 \\  
& Encoder\_10 & 0.022 & 0.046 & 0.276 & 0.732 \\  
& Decoder\_11 & 0.022 & 0.046 & 0.275 & 0.738 \\  
& Encoder\_11 & 0.022 & 0.047 & 0.273 & 0.747 \\  
& Decoder\_12 & 0.022 & 0.046 & 0.27 & 0.738 \\  
& Encoder\_12 & 0.022 & 0.046 & 0.273 & 0.738 \\  
& Decoder\_13 & 0.022 & 0.046 & 0.272 & 0.742 \\  
& Encoder\_13 & 0.022 & 0.046 & 0.27 & 0.739 \\  
& Decoder\_14 & 0.022 & 0.046 & 0.274 & 0.744 \\  
& Encoder\_14 & 0.022 & 0.046 & 0.272 & 0.735 \\  
& Decoder\_15 & 0.022 & 0.046 & 0.272 & 0.738 \\  
& Encoder\_15 & 0.022 & 0.046 & 0.274 & 0.737 \\  
& Encoder\_16 & 0.022 & 0.046 & 0.273 & 0.732 \\  
& None & 0.022 & 0.046 & 0.272 & 0.737 \\  
& Embedding & 0.045 & 0.062 & 0.652 & 0.955 \\ 
\bottomrule
\end{tabularx}
\end{table}

\begin{table}[H]
\caption{Ablation Experiment Results for Temperature and Charging Capacity}
\smallskip
\label{tabS2}
\begin{tabularx}{\textwidth}{YYYYYY}
\toprule
\textbf{Used Early Cycles} & \textbf{Detach} & \textbf{MAE\_T ($^\circ$C)} & \textbf{RMSE\_T ($^\circ$C)} & \textbf{MAE\_Qc (Ah)} & \textbf{RMSE\_Qc (Ah)} \\
\midrule 
\multirow{22}{*}{20} & Decoder\_1 & 0.402 & 0.768 & 0.02 & 0.038 \\  
& Decoder\_2 & 0.401 & 0.763 & 0.02 & 0.039 \\  
& Encoder\_2 & 0.419 & 0.78 & 0.021 & 0.04 \\  
& Decoder\_3 & 0.406 & 0.763 & 0.021 & 0.04 \\  
& Encoder\_3 & 0.408 & 0.771 & 0.02 & 0.039 \\  
& Decoder\_4 & 0.402 & 0.77 & 0.019 & 0.038 \\  
& Encoder\_4 & 0.411 & 0.77 & 0.022 & 0.043 \\  
& Decoder\_5 & 0.418 & 0.778 & 0.021 & 0.039 \\  
& Encoder\_5 & 0.4 & 0.76 & 0.02 & 0.039 \\  
& Decoder\_6 & 0.404 & 0.763 & 0.02 & 0.039 \\  
& Encoder\_6 & 0.405 & 0.764 & 0.02 & 0.039 \\  
& Decoder\_7 & 0.401 & 0.766 & 0.019 & 0.038 \\  
& Encoder\_7 & 0.415 & 0.775 & 0.02 & 0.039 \\  
& Decoder\_8 & 0.407 & 0.773 & 0.02 & 0.038 \\  
& Encoder\_8 & 0.402 & 0.767 & 0.02 & 0.038 \\  
& Decoder\_9 & 0.404 & 0.768 & 0.02 & 0.039 \\  
& Encoder\_9 & 0.399 & 0.768 & 0.02 & 0.038 \\  
& Decoder\_10 & 0.408 & 0.763 & 0.021 & 0.039 \\  
& Encoder\_10 & 0.402 & 0.766 & 0.02 & 0.039 \\  
& Decoder\_11 & 0.398 & 0.76 & 0.019 & 0.038 \\  
& Encoder\_11 & 0.398 & 0.761 & 0.02 & 0.039 \\  
& Decoder\_12 & 0.407 & 0.763 & 0.02 & 0.039 \\  
\bottomrule
\end{tabularx}
\end{table}

\begin{table}[H]
\ContinuedFloat 
\captionsetup{labelsep=period}
\caption{{\em Cont.}}
\begin{tabularx}{\textwidth}{YYYYYY}
\toprule
\textbf{Used Early Cycles} & \textbf{Detach} & \textbf{MAE\_T ($^\circ$C)} & \textbf{RMSE\_T ($^\circ$C)} & \textbf{MAE\_Qc (Ah)} & \textbf{RMSE\_Qc (Ah)} \\
\midrule
\multirow{10}{*}{20} & Encoder\_12 & 0.403 & 0.771 & 0.02 & 0.039 \\  
& Decoder\_13 & 0.403 & 0.762 & 0.02 & 0.038 \\  
& Encoder\_13 & 0.417 & 0.775 & 0.02 & 0.039 \\  
& Decoder\_14 & 0.415 & 0.763 & 0.021 & 0.038 \\  
& Encoder\_14 & 0.411 & 0.764 & 0.021 & 0.039 \\  
& Decoder\_15 & 0.406 & 0.77 & 0.02 & 0.038 \\  
& Encoder\_15 & 0.411 & 0.765 & 0.021 & 0.039 \\  
& Encoder\_16 & 0.411 & 0.765 & 0.021 & 0.039 \\  
& None & 0.404 & 0.753 & 0.021 & 0.039 \\  
& Embedding & 1.106 & 1.472 & 0.065 & 0.114 \\ 
\midrule
\multirow{32}{*}{40} & Decoder\_1 & 0.407 & 0.773 & 0.02 & 0.042 \\  
& Decoder\_2 & 0.398 & 0.762 & 0.02 & 0.041 \\  
& Encoder\_2 & 0.397 & 0.757 & 0.02 & 0.041 \\  
& Decoder\_3 & 0.407 & 0.763 & 0.021 & 0.042 \\  
& Encoder\_3 & 0.399 & 0.762 & 0.02 & 0.041 \\  
& Decoder\_4 & 0.399 & 0.757 & 0.02 & 0.042 \\  
& Encoder\_4 & 0.402 & 0.765 & 0.02 & 0.042 \\  
& Decoder\_5 & 0.402 & 0.762 & 0.02 & 0.042 \\  
& Encoder\_5 & 0.407 & 0.765 & 0.02 & 0.042 \\  
& Decoder\_6 & 0.402 & 0.764 & 0.02 & 0.041 \\  
& Encoder\_6 & 0.399 & 0.76 & 0.02 & 0.042 \\  
& Decoder\_7 & 0.402 & 0.756 & 0.021 & 0.042 \\  
& Encoder\_7 & 0.404 & 0.767 & 0.021 & 0.042 \\  
& Decoder\_8 & 0.402 & 0.766 & 0.02 & 0.042 \\  
& Encoder\_8 & 0.4 & 0.761 & 0.02 & 0.041 \\  
& Decoder\_9 & 0.402 & 0.757 & 0.02 & 0.041 \\  
& Encoder\_9 & 0.406 & 0.765 & 0.02 & 0.042 \\  
& Decoder\_10 & 0.405 & 0.76 & 0.021 & 0.042 \\  
& Encoder\_10 & 0.408 & 0.759 & 0.021 & 0.042 \\  
& Decoder\_11 & 0.402 & 0.762 & 0.02 & 0.042 \\  
& Encoder\_11 & 0.397 & 0.755 & 0.02 & 0.042 \\  
& Decoder\_12 & 0.402 & 0.76 & 0.02 & 0.042 \\  
& Encoder\_12 & 0.4 & 0.751 & 0.021 & 0.042 \\  
& Decoder\_13 & 0.406 & 0.768 & 0.021 & 0.042 \\  
& Encoder\_13 & 0.4 & 0.757 & 0.021 & 0.041 \\  
& Decoder\_14 & 0.398 & 0.755 & 0.02 & 0.041 \\  
& Encoder\_14 & 0.403 & 0.767 & 0.02 & 0.042 \\  
& Decoder\_15 & 0.4 & 0.758 & 0.02 & 0.041 \\  
& Encoder\_15 & 0.407 & 0.767 & 0.02 & 0.042 \\  
& Encoder\_16 & 0.405 & 0.761 & 0.021 & 0.043 \\  
& None & 0.41 & 0.769 & 0.021 & 0.042 \\  
& Embedding & 0.872 & 1.145 & 0.052 & 0.082 \\ 
\midrule
\multirow{12}{*}{60} & Decoder\_1 & 0.422 & 0.796 & 0.021 & 0.044 \\  
& Decoder\_2 & 0.427 & 0.801 & 0.021 & 0.044 \\  
& Encoder\_2 & 0.418 & 0.792 & 0.02 & 0.044 \\  
& Decoder\_3 & 0.427 & 0.804 & 0.021 & 0.044 \\  
& Encoder\_3 & 0.416 & 0.792 & 0.02 & 0.044 \\  
& Decoder\_4 & 0.427 & 0.8 & 0.021 & 0.045 \\  
& Encoder\_4 & 0.433 & 0.813 & 0.021 & 0.045 \\  
& Decoder\_5 & 0.424 & 0.806 & 0.02 & 0.044 \\  
& Encoder\_5 & 0.42 & 0.8 & 0.02 & 0.044 \\  
& Decoder\_6 & 0.421 & 0.795 & 0.02 & 0.044 \\  
& Encoder\_6 & 0.427 & 0.806 & 0.02 & 0.044 \\  
& Decoder\_7 & 0.417 & 0.786 & 0.02 & 0.044 \\  
\bottomrule
\end{tabularx}
\end{table}

\begin{table}[H]
\ContinuedFloat 
\captionsetup{labelsep=period}
\caption{{\em Cont.}}
\begin{tabularx}{\textwidth}{YYYYYY}
\toprule
\textbf{Used Early Cycles} & \textbf{Detach} & \textbf{MAE\_T ($^\circ$C)} & \textbf{RMSE\_T ($^\circ$C)} & \textbf{MAE\_Qc (Ah)} & \textbf{RMSE\_Qc (Ah)} \\
\midrule
\multirow{20}{*}{60} & Encoder\_7 & 0.419 & 0.794 & 0.02 & 0.044 \\  
& Decoder\_8 & 0.421 & 0.79 & 0.02 & 0.044 \\  
& Encoder\_8 & 0.429 & 0.805 & 0.021 & 0.045 \\  
& Decoder\_9 & 0.425 & 0.795 & 0.021 & 0.044 \\  
& Encoder\_9 & 0.422 & 0.794 & 0.02 & 0.044 \\  
& Decoder\_10 & 0.416 & 0.787 & 0.02 & 0.043 \\  
& Encoder\_10 & 0.418 & 0.794 & 0.021 & 0.044 \\  
& Decoder\_11 & 0.423 & 0.791 & 0.02 & 0.044 \\  
& Encoder\_11 & 0.424 & 0.793 & 0.02 & 0.044 \\  
& Decoder\_12 & 0.432 & 0.809 & 0.021 & 0.045 \\  
& Encoder\_12 & 0.422 & 0.809 & 0.02 & 0.044 \\  
& Decoder\_13 & 0.435 & 0.809 & 0.021 & 0.045 \\  
& Encoder\_13 & 0.418 & 0.796 & 0.02 & 0.044 \\  
& Decoder\_14 & 0.423 & 0.792 & 0.021 & 0.045 \\  
& Encoder\_14 & 0.419 & 0.793 & 0.02 & 0.044 \\  
& Decoder\_15 & 0.425 & 0.786 & 0.02 & 0.044 \\  
& Encoder\_15 & 0.423 & 0.8 & 0.02 & 0.044 \\  
& Encoder\_16 & 0.415 & 0.783 & 0.02 & 0.044 \\  
& None & 0.417 & 0.789 & 0.02 & 0.044 \\  
& Embedding & 1.083 & 1.454 & 0.067 & 0.115 \\  
\midrule
\multirow{32}{*}{80} & Decoder\_1 & 0.408 & 0.792 & 0.019 & 0.043 \\  
& Decoder\_2 & 0.407 & 0.789 & 0.02 & 0.044 \\  
& Encoder\_2 & 0.407 & 0.788 & 0.02 & 0.043 \\  
& Decoder\_3 & 0.407 & 0.781 & 0.019 & 0.043 \\  
& Encoder\_3 & 0.403 & 0.776 & 0.019 & 0.043 \\  
& Decoder\_4 & 0.407 & 0.778 & 0.02 & 0.043 \\  
& Encoder\_4 & 0.404 & 0.779 & 0.019 & 0.043 \\  
& Decoder\_5 & 0.405 & 0.787 & 0.019 & 0.043 \\  
& Encoder\_5 & 0.409 & 0.796 & 0.019 & 0.043 \\  
& Decoder\_6 & 0.403 & 0.788 & 0.019 & 0.043 \\  
& Encoder\_6 & 0.404 & 0.776 & 0.02 & 0.043 \\  
& Decoder\_7 & 0.406 & 0.788 & 0.019 & 0.043 \\  
& Encoder\_7 & 0.408 & 0.796 & 0.019 & 0.043 \\  
& Decoder\_8 & 0.405 & 0.789 & 0.019 & 0.043 \\  
& Encoder\_8 & 0.408 & 0.783 & 0.02 & 0.043 \\  
& Decoder\_9 & 0.405 & 0.783 & 0.019 & 0.043 \\  
& Encoder\_9 & 0.413 & 0.79 & 0.02 & 0.043 \\  
& Decoder\_10 & 0.406 & 0.781 & 0.019 & 0.043 \\  
& Encoder\_10 & 0.404 & 0.785 & 0.019 & 0.043 \\  
& Decoder\_11 & 0.41 & 0.781 & 0.02 & 0.043 \\  
& Encoder\_11 & 0.409 & 0.785 & 0.02 & 0.043 \\  
& Decoder\_12 & 0.406 & 0.781 & 0.019 & 0.043 \\  
& Encoder\_12 & 0.41 & 0.78 & 0.019 & 0.043 \\  
& Decoder\_13 & 0.409 & 0.784 & 0.019 & 0.043 \\  
& Encoder\_13 & 0.406 & 0.783 & 0.02 & 0.043 \\  
& Decoder\_14 & 0.404 & 0.783 & 0.019 & 0.043 \\  
& Encoder\_14 & 0.405 & 0.773 & 0.019 & 0.043 \\  
& Decoder\_15 & 0.406 & 0.782 & 0.019 & 0.043 \\  
& Encoder\_15 & 0.401 & 0.778 & 0.019 & 0.043 \\  
& Encoder\_16 & 0.409 & 0.794 & 0.019 & 0.043 \\  
& None & 0.405 & 0.787 & 0.019 & 0.043 \\  
& Embedding & 0.924 & 1.203 & 0.053 & 0.082 \\  
\bottomrule
\end{tabularx}
\end{table}

\begin{table}[H]
\ContinuedFloat 
\captionsetup{labelsep=period}
\caption{{\em Cont.}}
\begin{tabularx}{\textwidth}{YYYYYY}
\toprule
\textbf{Used Early Cycles} & \textbf{Detach} & \textbf{MAE\_T ($^\circ$C)} & \textbf{RMSE\_T ($^\circ$C)} & \textbf{MAE\_Qc (Ah)} & \textbf{RMSE\_Qc (Ah)} \\
\midrule
\multirow{32}{*}{100} & Decoder\_1 & 0.409 & 0.79 & 0.02 & 0.044 \\  
& Decoder\_2 & 0.4 & 0.78 & 0.02 & 0.043 \\  
& Encoder\_2 & 0.402 & 0.769 & 0.02 & 0.044 \\  
& Decoder\_3 & 0.407 & 0.783 & 0.02 & 0.043 \\  
& Encoder\_3 & 0.406 & 0.777 & 0.02 & 0.044 \\  
& Decoder\_4 & 0.405 & 0.775 & 0.02 & 0.043 \\  
& Encoder\_4 & 0.399 & 0.765 & 0.02 & 0.043 \\  
& Decoder\_5 & 0.403 & 0.772 & 0.02 & 0.043 \\  
& Encoder\_5 & 0.399 & 0.776 & 0.02 & 0.043 \\  
& Decoder\_6 & 0.409 & 0.786 & 0.02 & 0.044 \\  
& Encoder\_6 & 0.4 & 0.767 & 0.02 & 0.043 \\  
& Decoder\_7 & 0.401 & 0.779 & 0.02 & 0.043 \\  
& Encoder\_7 & 0.406 & 0.775 & 0.02 & 0.043 \\  
& Decoder\_8 & 0.402 & 0.768 & 0.02 & 0.043 \\  
& Encoder\_8 & 0.404 & 0.79 & 0.02 & 0.044 \\  
& Decoder\_9 & 0.4 & 0.769 & 0.02 & 0.043 \\  
& Encoder\_9 & 0.403 & 0.784 & 0.02 & 0.044 \\  
& Decoder\_10 & 0.405 & 0.783 & 0.02 & 0.044 \\  
& Encoder\_10 & 0.405 & 0.774 & 0.02 & 0.043 \\  
& Decoder\_11 & 0.405 & 0.777 & 0.02 & 0.043 \\  
& Encoder\_11 & 0.406 & 0.783 & 0.02 & 0.044 \\  
& Decoder\_12 & 0.402 & 0.77 & 0.019 & 0.043 \\  
& Encoder\_12 & 0.404 & 0.779 & 0.02 & 0.043 \\  
& Decoder\_13 & 0.404 & 0.783 & 0.02 & 0.043 \\  
& Encoder\_13 & 0.4 & 0.776 & 0.02 & 0.044 \\  
& Decoder\_14 & 0.403 & 0.78 & 0.02 & 0.043 \\  
& Encoder\_14 & 0.401 & 0.767 & 0.02 & 0.043 \\  
& Decoder\_15 & 0.401 & 0.769 & 0.02 & 0.043 \\  
& Encoder\_15 & 0.404 & 0.777 & 0.021 & 0.044 \\  
& Encoder\_16 & 0.411 & 0.783 & 0.02 & 0.044 \\  
& None & 0.404 & 0.773 & 0.02 & 0.044 \\  
& Embedding & 0.921 & 1.216 & 0.053 & 0.085 \\  
\bottomrule
\end{tabularx}
\end{table}

\end{document}